\newcommand{\floatMargin}{-1.6mm}

\PassOptionsToPackage{pdfpagelabels=false}{hyperref}
\documentclass{sig-alternate}[10pt]
\usepackage[
  pass,
]{geometry}
\usepackage{comment}
\usepackage{mathptmx} 

\newcommand{\ignore}[1]{}
\usepackage{fancyhdr}
\usepackage[normalem]{ulem}
\usepackage[hyphens]{url}

\usepackage{amsmath}
\usepackage{graphicx}
\usepackage{xcolor}
\usepackage{xspace}
\usepackage{bbold}
\usepackage{pbox}
\usepackage[rightcaption]{sidecap}
\usepackage{flushend}
\sidecaptionvpos{figure}{c}

\usepackage[caption=false]{subfig}
\usepackage{hyperref}
\usepackage{soul}
\usepackage{enumitem}

\renewcommand\hl[1]{#1}

\usepackage{fixltx2e}

\definecolor{Awesome}{rgb}{1.0, 0.08, 0.58}

\newcommand{\DPNN}{\textit{BASE}\xspace}

\newcommand{\BASE}{BASE\xspace}

\newcommand{\DPRL}{\textit{Dynamic Precision Reduction}\xspace}

\newcommand{\DPRTITLE}{DPRed\xspace}
\newcommand{\DPR}{\textit{DPRed}\xspace}

\newcommand{\STRL}{\textit{Stripes}\xspace}
\newcommand{\STRLTITLE}{STRIPES\xspace}
\newcommand{\STR}{\textit{Stripes}\xspace}
\newcommand{\STRTITLE}{{STRIPES}\xspace}

\newcommand{\TRTL}{\textit{Tartan}\xspace}
\newcommand{\TRTLTITLE}{TARTAN\xspace}
\newcommand{\TRT}{\textit{TRT}\xspace}
\newcommand{\TRTTITLE}{TRT\xspace}
\newcommand{\DSTRL}{\textit{\DPR Stripes}\xspace}
\newcommand{\DSTRLTITLE}{\DPRTITLE Stripes\xspace}
\newcommand{\DSTR}{\textit{DStripes}\xspace}
\newcommand{\DSTRTITLE}{{DStripes}\xspace}
\newcommand{\TRTEVAL}{\textit{DStripes\textsubscript{T}}\xspace}

\usepackage{xstring}

\makeatletter
\def\@separatefourdigits#1#2{%
  \StrLen{#1}[\length]%
  \ifnum\length<5\relax#1\else%
     \bgroup
     \StrRight{#1}{4}[\lowerfour]%
     \StrGobbleRight{#1}{4}[\remaining]%
     \@separatefourdigits{\remaining}{#2}#2\lowerfour%
     \egroup
  \fi}

\def\formatnumber#1#2{\StrDel{#1}{ }[\aux]%
\@separatefourdigits{\aux}{#2}%
}

\makeatother

\title{\DPRTITLE: Making Typical Activation and Weight Values Matter In Deep Learning Computing}

\author{Alberto Delm\'as Lascorz, Sayeh Sharify, Patrick Judd, Kevin Siu, Milos Nikolic, Andreas Moshovos\\
Department of Electrical and Computer Engineering, University of Toronto\\ 
\texttt{\{delmasl1, sayeh, juddpatr, moshovos\}@ece.utoronto.ca}\\ 
\texttt{\{kcm.siu, milos.nikolic\}@mail.utoronto.ca}
\\ }

\newcommand\blfootnote[1]{%
  \begingroup
  \renewcommand\thefootnote{}\footnotetext{#1}%
  \addtocounter{footnote}{-1}%
  \endgroup
}

\begin{document}
\maketitle

\newcommand{\meanspeedupFC}{$1.52\times$\xspace}
\newcommand{\meanspeedupCV}{$2.65\times$\xspace}
\newcommand{\meanspeedupALL}{$2.59\times$\xspace} 

\newcommand{\meanefficiencyFC}{$1.05\times$\xspace}
\newcommand{\meanefficiencyCV}{$1.21\times$\xspace}
\newcommand{\meanefficiencyALL}{$1.19\times$\xspace} 

\newcommand{\meanspeedupFCloss}{$1.82\times$\xspace}
\newcommand{\meanspeedupCVloss}{$3.09\times$\xspace}
\newcommand{\meanspeedupALLloss}{$3.03\times$\xspace} 

\newcommand{\meanefficiencyFCloss}{$1.25\times$\xspace}
\newcommand{\meanefficiencyCVloss}{$1.41\times$\xspace}
\newcommand{\meanefficiencyALLloss}{$1.40\times$\xspace}

\begin{abstract}
We show that selecting a single data type (precision) for \textit{all} values in Deep Neural Networks, even if that data type is different per layer, amounts to \textit{worst case} design. Much shorter data types can be used if we target the \textit{common case} by adjusting the precision at a much finer granularity.  We propose \DPRL (\DPR), where we group weights and activations and encode them using a precision specific to each group. The per group precisions are selected statically for the weights and dynamically by hardware for the activations. We exploit these precisions to reduce: 1) off-chip storage and off- and on-chip communication, and 2) execution time. \DPR compression reduces off-chip traffic to nearly 35\% and \hl{33\%} on average compared to no compression respectively for 16b \hl{and 8b models}. This makes it possible to sustain higher performance for a given off-chip memory interface while also boosting energy efficiency. We also demonstrate designs where the time required to process each group of activations and/or weights scales proportionally to the precision they use for convolutional and fully-connected layers. This improves execution time and energy efficiency for both dense and sparse networks. 
\hl{We show the techniques work with 8-bit networks, where 1.82x and 2.81x speedups are achieved for two different hardware variants that take advantage of dynamic precision variability.}

\end{abstract}

\section{Introduction}

Early successes in hardware acceleration for Deep Learning Neural Network (DNN)  relied on exploiting its computation structure and the reuse in its access stream, e.g.,~\cite{diannao,DaDiannao,isscc_2016_chen_eyeriss}. Followup work identified and exploited various forms of \textit{informational inefficiency} in DNNs such as ineffectual neurons~\cite{han_eie:_2016,SCNN}, activations~\cite{han_eie:_2016,albericio:cnvlutin,SCNN}, ineffectual weights~\cite{CambriconXMICRO16,SCNN},  an excess of precision~\cite{kim_x1000_2014,judd:reduced,binaryconnect,quantizedBlog,DBLP:journals/corr/VenkateshNM16,Stripes-MICRO,bitfusion,outlier},  ineffectual activation bits~\cite{Pragmatic}, and hyper-parameter over-provisioning, e.g.,~\cite{squeezenet,pruneornot}. Identifying additional forms of informational inefficiency is invaluable as it opens up additional opportunities for boosting execution time performance and energy efficiency which in turn support further innovation in DNN applications and design.

\blfootnote{This paper was submitted for review (August 3, 2018, revised on October 19, 2018) to the IEEE International Conference on High-Performance Computer Architecture 2019. 
Earlier versions were submitted for review to the:
 2018 ACM/IEEE International Symposium on Microarchitecture (April 6, 2018), 
the 2018 ACM/IEEE International Symposium on Computer Architecture and the 2018 (November 21, 2017),  IEEE International Conference on High Performance Computer Architecture (August 1, 2017). A preliminary evaluation of dynamic precision adaptation was presented by Delm\'as Lascorz \textit{et al~\cite{DBLP:journals/corr/DelmasJSM17}.}}

We highlight an overlooked informational inefficiency in the value stream of DNNs: the datatype/precision needed by most activations and weights is much shorter than that needed when considering the network as a whole or each of its layers individually. Additionally, the datatype needed by each activation varies considerably with the specific \textit{runtime} input. In retrospect, the above observations are unsurprising: 1)~by design, the expected per layer distribution of values, be it for weights or activations, is that most will be near zero and few will be of higher magnitude, and 2)~the runtime values will obviously depend on the input. 

\sloppy
Surprisingly, however, no hardware to date exploits the fine-grain precision variability of weights and activations fully. Some designs exploit \textit{profiled derived} or \textit{quantized} per layer precisions by: hardwiring the precision~per layer~\cite{kim_x1000_2014}, scaling voltage and frequency per layer~\cite{envision}, or supporting several data widths, e.g., ~\cite{TPUISCA17,bitfusion} or the full spectrum of bit-widths~\cite{Stripes-MICRO}.  These statically chosen per layer precisions must accommodate \textit{any possible input} and for \textit{any} activation and weight across a \textit{whole} layer.  We highlight that this design corresponds to a \textit{worst case} precision analysis that exacerbates the importance of the exceedingly rare activation and weight values of large magnitude. 
We show that much higher potential for improvement exists if we tailor precision to target the \textit{common case} instead. Specifically, the precision used at any given point of time needs to accommodate: 1)~the activation values for the \textit{specific} input at hand, and further 2)~only the activation and weight values that are being  processed concurrently.  As a result the precision can vary with the input and could be adapted at a much finer granularity than the layer. 
 
\sloppy 
We demonstrate that, depending on how many activations and weights are processed together, the precision that they need can be lower than that identified through per layer profiling. We propose \textit{Dynamic Precision Reduction} or \DPR (pronounced ``Deep Red'') which adapts precision at a fine-granularity by grouping activations and weights and choosing a precision for each group separately. \DPR chooses precisions statically for the weights and dynamically for the activations.   An accelerator that incorporates \DPR can use it to improve performance, reduce communication and storage needs, and ultimately improve energy efficiency over one that merely uses profile-derived per layer precisions.

\noindent We make the  following contributions:
\\
\textbf{1. }We characterize the degree in which  the data type needed by the weights and activations in DNNs varies when using groups much smaller than the layer and with the input.  \\
\sloppy
\textbf{2.} We propose a DPRed hardware building block for compressing and decompressing groups of weights and activations off-chip thus reducing energy considerably and boosting the effective off-chip bandwidth. For weights the compression is done once as a pre-processing step. For activations, it is performed dynamically at the output of the previous layer. In both cases, the compressed data is decompressed on-the-fly when fetched from off-chip. This compression scheme reduces off-chip traffic to 38\%. By comparison, using a fixed, profile precision reduces off-chip traffic to 50\%. This building block is general and can be used with numerous accelerators.\\
\sloppy
\textbf{3.} We present \DSTRL (\DSTR), an accelerator whose performance, communication and storage vary proportionally with the precision of activations at the granularity of a processing group. \DSTR builds upon the \STRL accelerator~\cite{Stripes-MICRO} which exploits per-layer profile-derived precisions. A major advantage of \DSTR is that the hardware changes it needs are modest yet the resulting performance, communication, storage, and energy efficiency improvements are anything but. For example, compared to a fixed-precision accelerator \DPNN (see Section~\ref{sec:baseline}) and for a configuration that performs 4K $16b\times 16b$ multiplications per cycle,  \STR improves average performance and energy efficiency assuming everything fits on chip by $1.9\times$ and $1.32\times$ while \DSTR improves them by $2.6\times$ and $1.84\times$ respectively. 
\\ 
\textbf{4.} We show that \DSTR can also deliver benefits for pruned models (AlexNet, GoogleNet, ResNet50, and MobileNet) and that it delivers higher performance (on average $3.5\times$) than a similarly configured SCNN~\cite{SCNN} (on average $1.9\times$). Moreover, we explain that \DPR is compatible with SCNN, and could be useful when processing larger inputs such as high-resolution images.
\\
\textbf{5. } Like \STRL, \DSTR accelerates computation only for convolutional layers which dominate execution time in many image related applications~\cite{DaDiannao}. Since fully-connected layers are more prevalent in other applications~\cite{TPUISCA17} we present  
\textit{TARTAN} (\TRT), which enables \DSTR to exploit precision variability also for these layers too; \TRT is $P/{max(P_w,P_a)}$ faster than \STRL and \BASE where $P$ is the maximum precision supported in hardware, and $P_w$ and $P_a$ are the per group precisions for the weights and the activations respectively. \TRT can be used independently of dynamic precision reduction. When combined with \DSTR, \TRT improves performance over \DPNN by \meanspeedupALL and \meanefficiencyALL over the fixed-precision accelerator for a \textit{broader} set of neural networks. 

In 65nm \DSTR requires less than 1\% more area than \STRL. Adding \TRT requires 26-51\% more area depending on its configuration. Improvements with \DSTR alone, or with \DSTR combined with \TRT far exceed those possible by scaling \DPNN to use the same area.
\\
\textbf{6.} \hl{Since quantization has become more prevalent for certain classes of Deep Learning applications, we show that }\DPR \hl{can deliver benefits even for 8b quantized networks. We show memory traffic reduction, performance and energy improvements for 8b configurations of } \DSTR
\hl{and LOOM}~\cite{Sharify:2018:LEW:3195970.3196072},\hl{ a bit-serial design that exploits precisions for both weights and activations.}

\newcommand{\mytrim}{width=.3\linewidth}
\newcommand{\myscale}{.3}

\section{Dynamic Precision Variability}
\label{sec:dpr}

\begin{figure*}[t!] 
\centering
\subfloat[GoogLeNet, Conv 1]{\includegraphics[width=\myscale\linewidth]{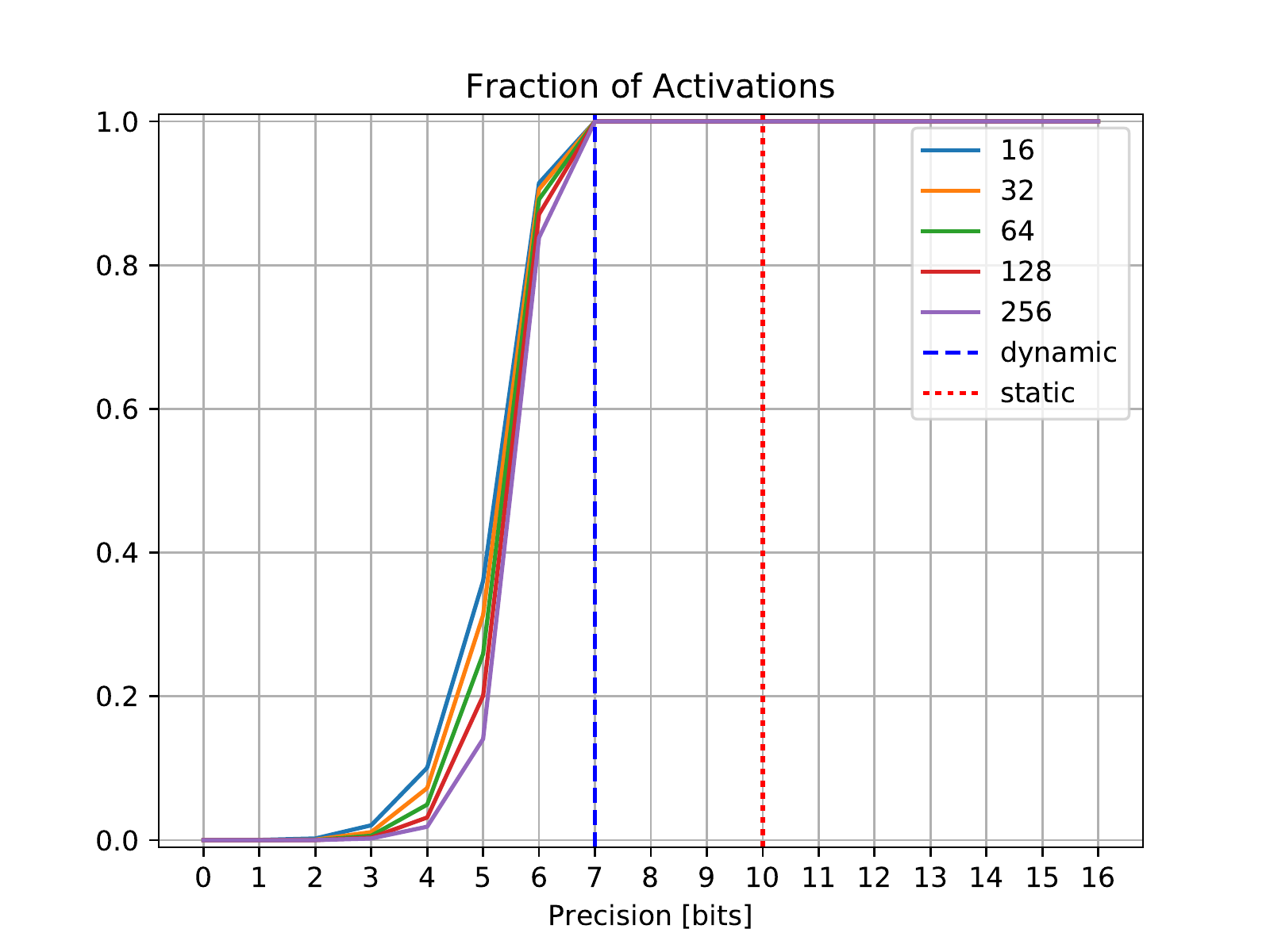}
\label{fig:google1}
}
\subfloat[GoogLeNet, 5a-1x1]{\includegraphics[width=\myscale\linewidth]{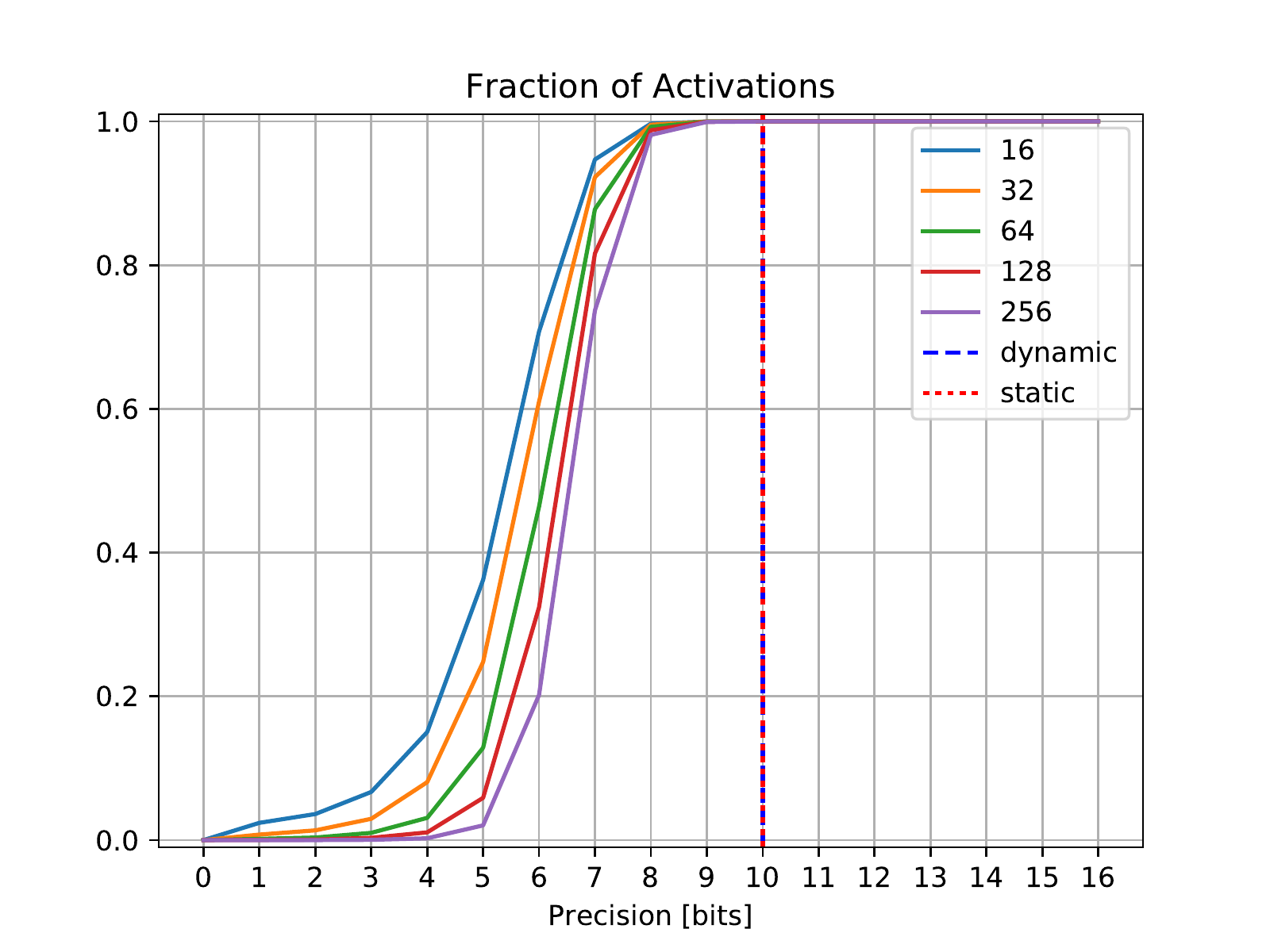}
\label{fig:google5a}
}
\vspace{-0.38cm}
\subfloat[SkimCaffe ResNet50, res3a\_branch1]{\includegraphics[width=\myscale\linewidth]{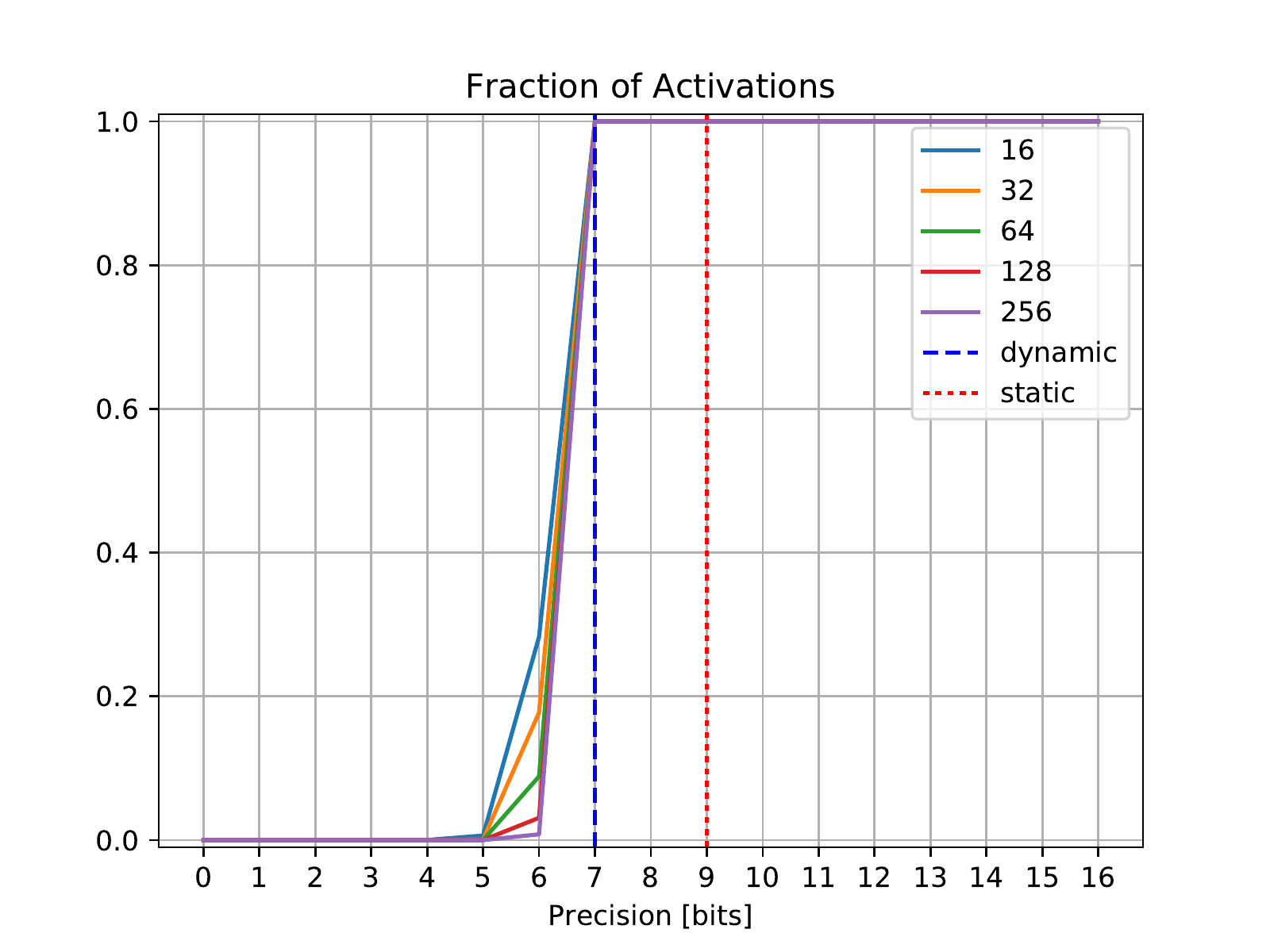}
\label{fig:resnet3}
}\\
\subfloat[SkimCaffe ResNet50, res5a\_branch1]{\includegraphics[width=\myscale\linewidth]{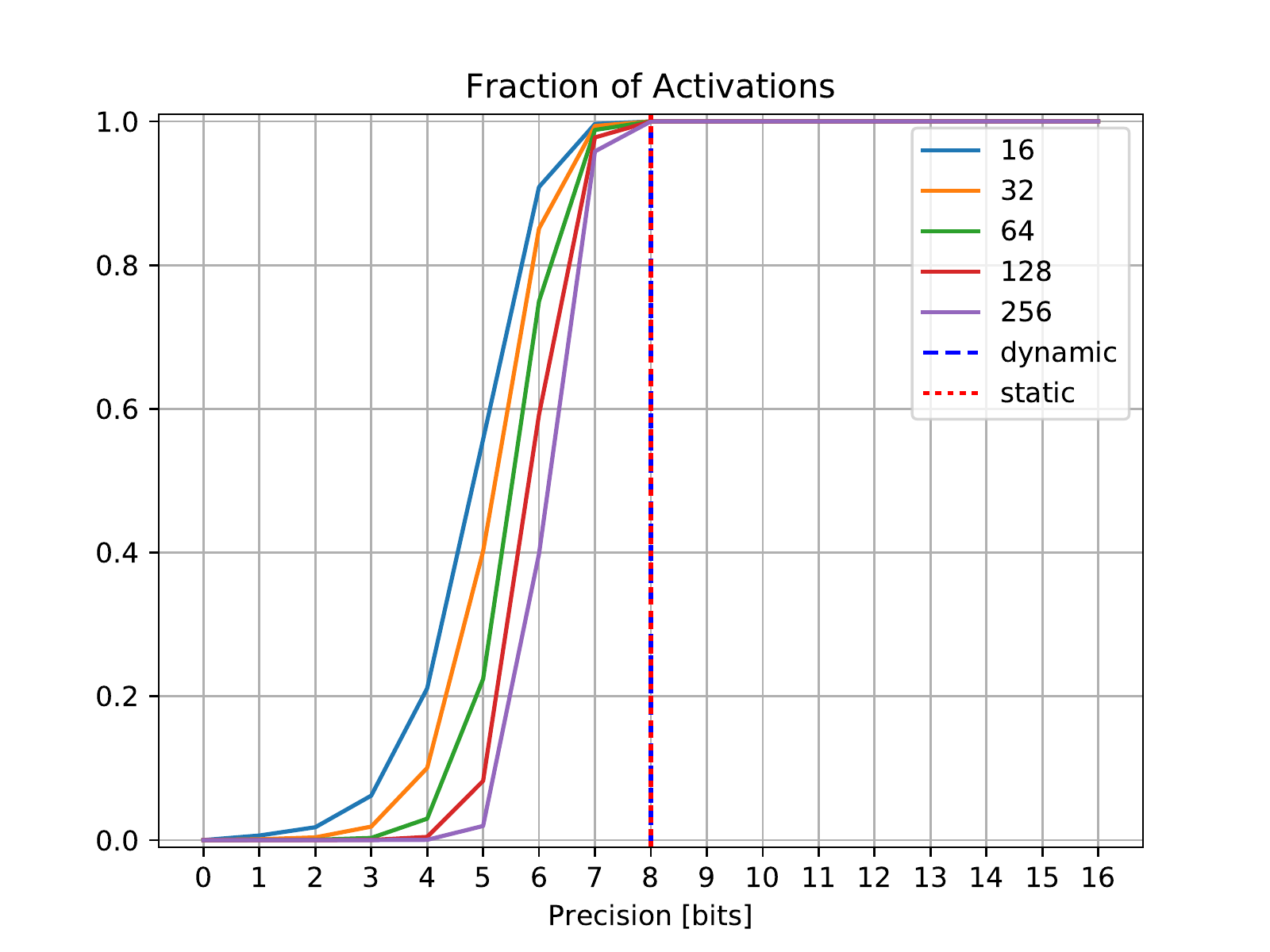}
\label{fig:resnet5}
}
\subfloat[{Other Networks}]{\includegraphics[width=\myscale\linewidth]{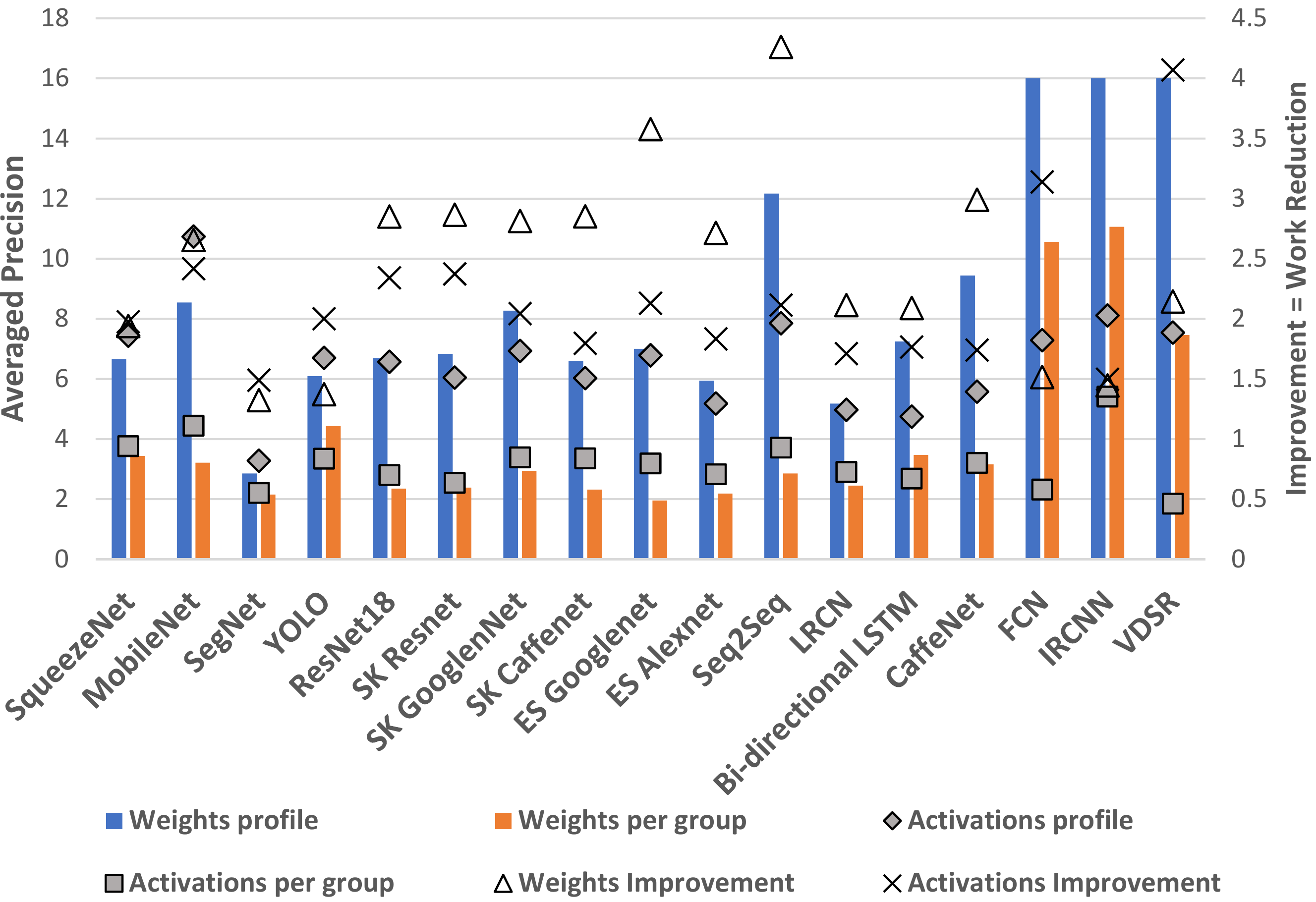}
\label{fig:othernets}
}
\subfloat[\hl{Quantized 8b GoogleNet}]{\includegraphics[width=\myscale\linewidth]{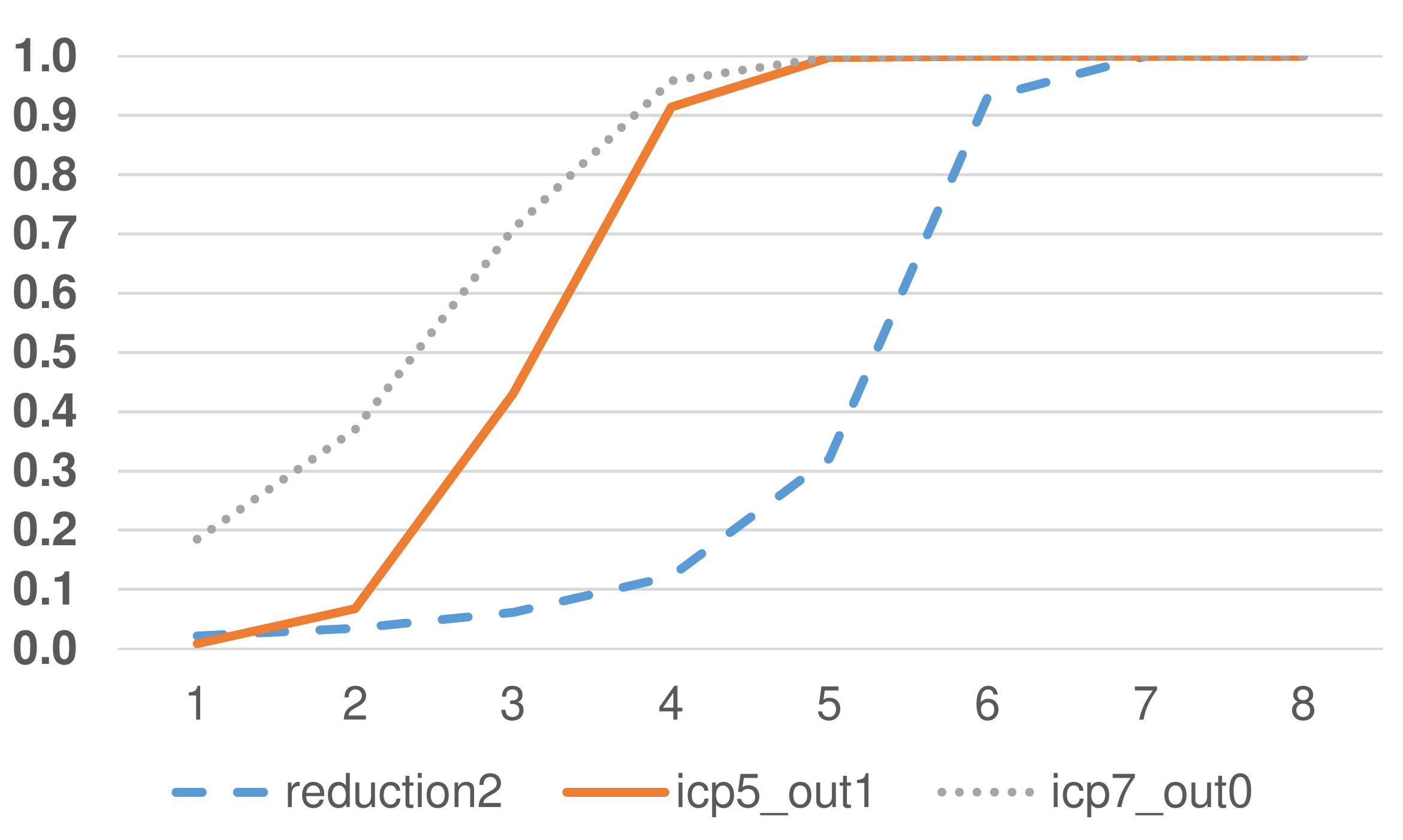}
\label{fig:google8b}
}
\caption{Precision profiles for some conv. layers. (a)-(d) Average over multiple images.  Multiple group sizes shown. "Dynamic" results for one sample image. 
{(e)~Average precisions and potential improvements for several networks. \hl{(f)~Dynamic precision variability for three layers of an 8b quantized version of GoogleNet.}}}
\label{fig:ppi}
\end{figure*}

This section demonstrates that using per layer precisions for the activations is overly pessimistic. We demonstrate benefits for weighs and activations in Section~\ref{sec:evaluation}.

\sloppy
Figures~\ref{fig:google1}-\ref{fig:resnet5} show precision measurements for four convolutional layers -- two from each of GoogleNet and SkimCaffe-ResNet50~\cite{SkimCaffePaper} a weight pruned network. Similar trends were observed in other networks and layers. The measurements are over 5,000 randomly selected images from the IMAGENET dataset~\cite{imagenet}. The graphs show the cumulative distribution of the precisions needed per group of activations for various group sizes ranging from 16 to 256, where each group uses the precision needed for its least favorable activation. Two vertical lines report the precisions when considering all activation values: 1)~the profile-derived (``static'') over all sample images, and 2)~the one that can be detected dynamically (``dynamic'') for one randomly selected sample image to illustrate that precisions also vary per input. 

Figure~\ref{fig:google1} reports measurements for \texttt{conv1}, the first layer of GoogleNet. The profile-determined precision is 10 bits  while for one specific image all values within the layer could be represented with just 7 bits, an improvement of 30\% in precision length.  The cumulative distribution of the precisions needed per group of 256 activations shows that further reduction in precision is possible. For example, about 80\% of these groups require 6 bits only, and about  15\% just 5 bits. Smaller group sizes further reduce the \textit{effective} precision, however, the differences are modest. These results suggest that picking a precision for the whole layer exacerbates the importance of a few high-magnitude activations.  

The first layer usually exhibits different value behavior than the rest since it processes the direct input image to discover visual features whereas the inner layers tend to try to find correlations among features. Layer \texttt{5a-1x1} in Figure~\ref{fig:google5a} shows more pronounced improvements in effective precision with the smaller group sizes. Figures~\ref{fig:resnet3} and~\ref{fig:resnet5} show that the behavior persists even in the sparse network. We include this measurement to demonstrate that dynamic precision variability is a phenomenon that is orthogonal to weight sparsity and thus of potential value to designs that target sparse neural networks.

{Figure}~\ref{fig:othernets} {compares (left Y-axis) the average effective precision with per layer profiling or with per value detection for weights and activations for additional models. These measurements are for \textit{1000} randomly selected images from IMAGENET}~\cite{imagenet} { for the image classification models}~\cite{caffe,MS152,SkimCaffePaper,cvpr_2017_yang_energy,squeezenet,MobileNets}{ and for 100 images from CamVid}~\cite{BrostowFC:PRL2008}{ for SegNet}~\cite{badrinarayanan2015segnet}{(segmentation), 100 and 10 inputs from Pascal VOC}~\cite{PascalVOC}{ respectively for YOLO V2}~\cite{DBLP:journals/corr/RedmonF16}{ (detection) and FCN8}~\cite{fcn}{ (segmentation), 10 images from CBSD68}~\cite{CBSD68_2,set5,set14}{, 10 inputs for IRCNN}~\cite{IRCNN}{ (denoising) and VDSR}~\cite{VDSR}{ (super-resolution), 500 inputs from WMT14}{ for Seq2Seq}~\cite{Seq2Seq}{ (translation), 500 inputs from COCO}~\cite{DBLP:journals/corr/LinMBHPRDZ14}{ for LRCN}~\cite{lrcn2014}{ (captioning), and 500 inputs from Flickr8k}~\cite{flickr8k} {for Bi-Directional LSTM}~\cite{bidirectionalLSTM}{ (captioning). The measurements are across the whole network. It also reports the reduction in work (right Y-axis) for weights and activations when per-value precision is used. The results illustrate that selecting a single precision per layer grossly overestimates the precision needed for the common case.} 

\hl{Figure}~\ref{fig:google8b} \hl{reports the distribution of precisions needed for an 8b quantized GoogleNet. Three of the longer running layers are shown which are representative of the range of behaviors seen over all layers. For }\texttt{reduction} \hl{and} \texttt{icp5\_out1}\hl{ more than 90\% of the activations can be represented with 4 bits or less. For }\texttt{icp7\_out0} \hl{more than 90\% of the activations need 6 bits a 25\% reduction over 8 bits.} Section~\ref{sec:quantized} \hl{shows that this behavior translates into significant memory and execution time benefits.}

Generally, for all the networks studied the following were observed: 1)~The activation precision needed varies, more so at the first layer, 2)~for all layers the precisions needed at a finer than a layer granularity are shorter than those needed for the whole layer, and 3)~only a small number of activation groups require the maximum precision needed for the layer as a whole. These results motivate incorporating \textit{dynamic precision reduction} in hardware accelerators.

\begin{table}
\centering
\caption{Average Per Layer Activation Precisions with DPRed.}
\label{tbl:effpre}
\scriptsize
\begin{tabular}{|l|l|l|}
\hline
\textbf{Network} & \textbf{Effective Precision Per Layer} & \textbf{Reduction}                                                 \\ \hline
\textbf{AlexNet} & 5.39-7.36-4.22-4.40-5.81                                                                & 22.59\% \\ \hline
\textbf{NiN} & 6.37-7.13-7.79-6.97-5.77-5.15-4.73-6.78&\\& -8.36-7.51-7.64-7.58                                                                & 23.56\% \\ \hline
\textbf{GoogleNet} & 6.19-5.94-5.74-6.77-6.91-6.77-6.86-6.77&\\& -6.92-6.31-5.96-6.31-6.00-6.31-6.55-5.33&\\& -5.33-5.33-5.33-5.33-5.48-6.74-6.33-6.74&\\& -6.51-6.74-7.07-6.35-6.17-6.35-5.88-6.35&\\& -6.56-5.07-4.69-5.07-4.82-5.07-5.31-5.53&\\& -4.89-5.53-5.70-5.53-5.86-7.88-7.62-7.88&\\& -8.07-7.88-8.31-4.97-3.85-4.97-3.61-4.97-5.36                                                                & 36.30\% \\ \hline
\textbf{VGG\_S}  & 5.28-5.05-5.82-3.38-4.80                                                                & 38.85\% \\ \hline
\textbf{VGG\_M}  & 5.28-5.05-5.04-5.37-4.00                                                                & 30.47\% \\ \hline
\textbf{VGG\_19} & 9.05-7.69-10.04-9.00-11.08-8.74-9.65-8.29-11.55&\\
&-10.37-12.22-11.67-11.53-11.54-10.40-5.9 & 21.20\% \\ \hline
\end{tabular}
\vspace{\floatMargin}
\end{table}

\begin{table}
\centering
\caption{Average Per Layer Weight Precisions with DPRed.}
\label{tbl:effprew}
\scriptsize
\begin{tabular}{|l|l|l|}
\hline
\textbf{Network} & \textbf{Effective Precision Per Layer}                                                  & \textbf{Reduction}\\ \hline
\textbf{AlexNet} & 8.36-7.62-7.62-7.44-7.55 & 22.55\%
\\ \hline 
\textbf{NiN} & 8.85-10.29-10.21-7.65-9.13-9.04-7.63-8.65&\\& -8.62-7.79-7.96-8.18 &  19.20\%                                                             \\ \hline
\textbf{GoogleNet} & 9.80-10.91-9.35-10.21-10.02-9.02-10.06&\\& -9.34-10.29-9.61-9.73-8.63-9.93-8.65-9.64&\\& -9.31-9.48-8.79-9.27-8.94-9.30-9.10-9.03&\\& -8.25-7.27-8.56-9.15-9.26-9.17-8.21-7.32&\\& -8.55-9.01-9.15-9.33-8.22-9.21-8.44-9.12&\\& -8.54-8.68-7.47-9.11-7.93-8.86-8.52-8.59&\\& -7.77-8.85-8.14-8.88-8.28-8.81-7.61-8.81&\\& -7.52-8.40 & 17.27\%                                                              \\ \hline
\textbf{VGG\_S}  & 9.94-6.96-8.53-8.13-8.10 & 22.26\%
\\ \hline 
\textbf{VGG\_M}  & 9.87-7.55-8.52-8.16-8.14 & 21.76\%
\\ \hline 
\textbf{VGG\_19} & 10.98-9.81-9.31-9.09-8.58-8.04-7.89-7.86& \\& -7.51-7.20-7.36-7.47-7.61-7.66-7.66-7.63 & 29.65\%
\\ \hline  
\end{tabular}
\vspace{\floatMargin}
\end{table}

\section{Reducing Effective Precision}
Tables~\ref{tbl:effpre} and~\ref{tbl:effprew} report the average precision DPRed achieves per layer which 
demonstrates that it can effectively reduce precisions. 
Due to space limitations we report results for a group size of 16 values along the channel dimension only. The effective precisions are fractional as they are averaged over all groups within the layer and weighted accordingly to their use at runtime.

\section{Reducing Off-Chip Storage and Communication}
The bulk of energy in DNNs is expended by off- and on-chip memory accesses. Further, when on-chip storage is limited, off-chip bandwidth can easily be the bottleneck and more easily so for the fully-connected layers due to limited data reuse. Fortunately, the variable precision needs of activations and weights can be used to  reduce the amount of storage and bandwidth needed on- and off-chip. Here we limit attention  to the off-chip compression scheme.  

\begin{figure*}[htb!] 
\centering
\subfloat[][Data Type Conversion]{
\includegraphics[scale=0.2]{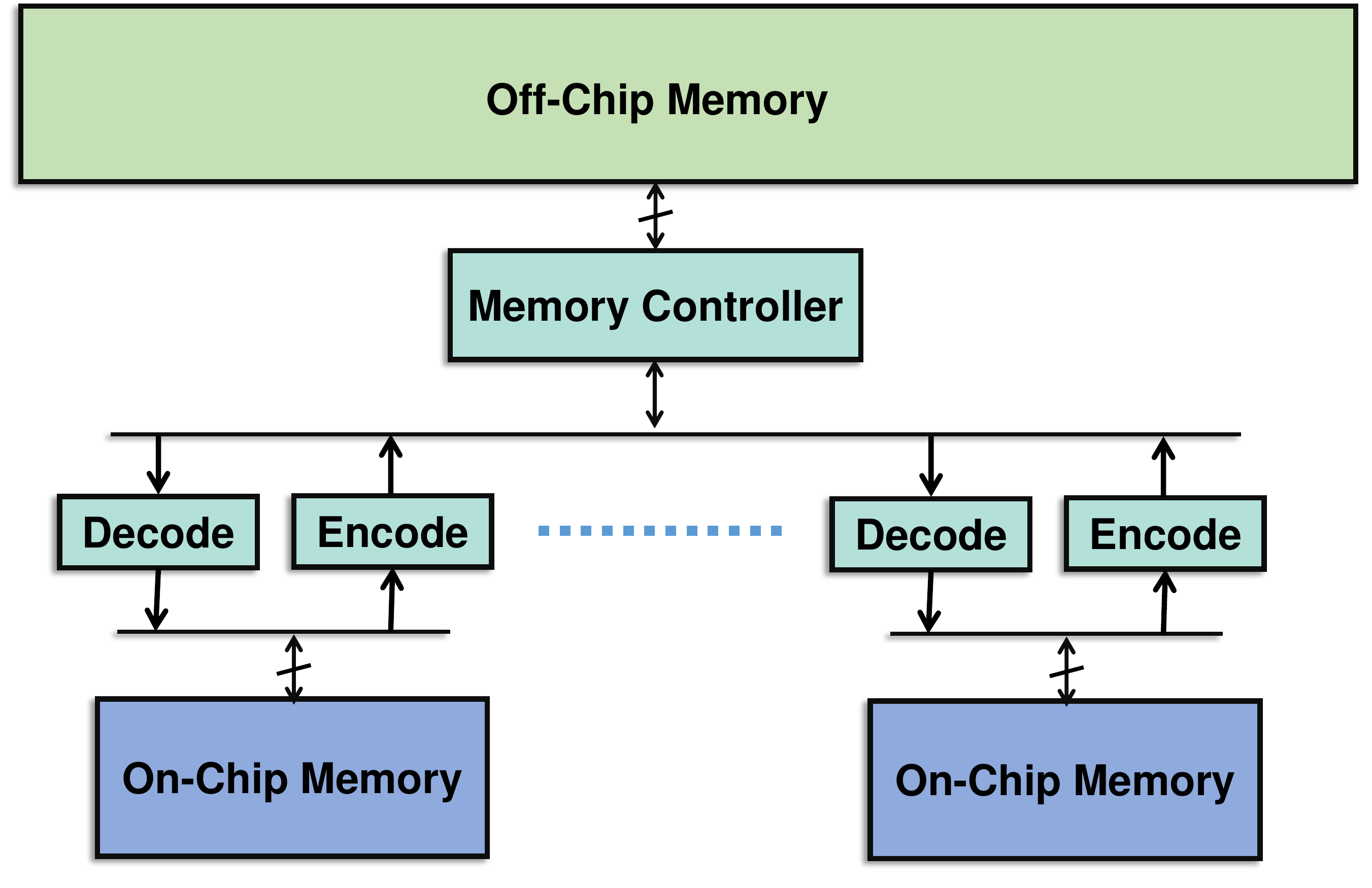}
\label{fig:memarch}
}
\subfloat[][Off-chip Group Data Container]{
\includegraphics[scale=0.2]{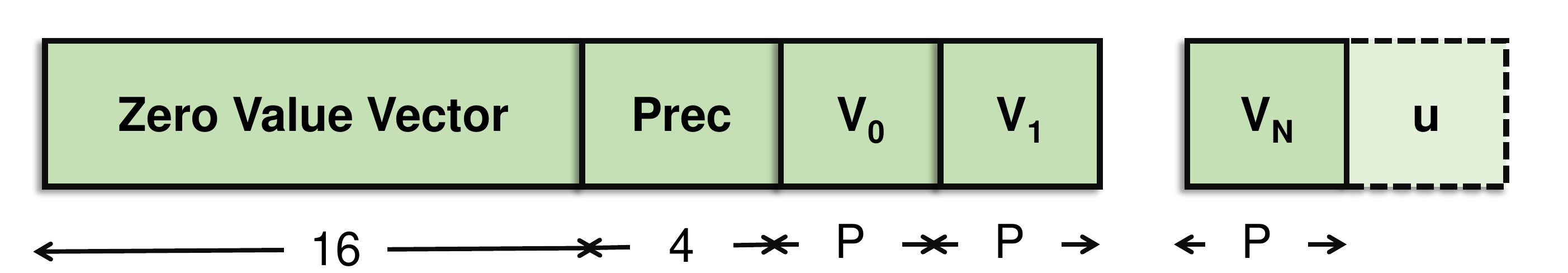}
\label{fig:memcontainer}
}
\subfloat[][Detecting per group precisions for activations.]{
\includegraphics[scale=0.25]{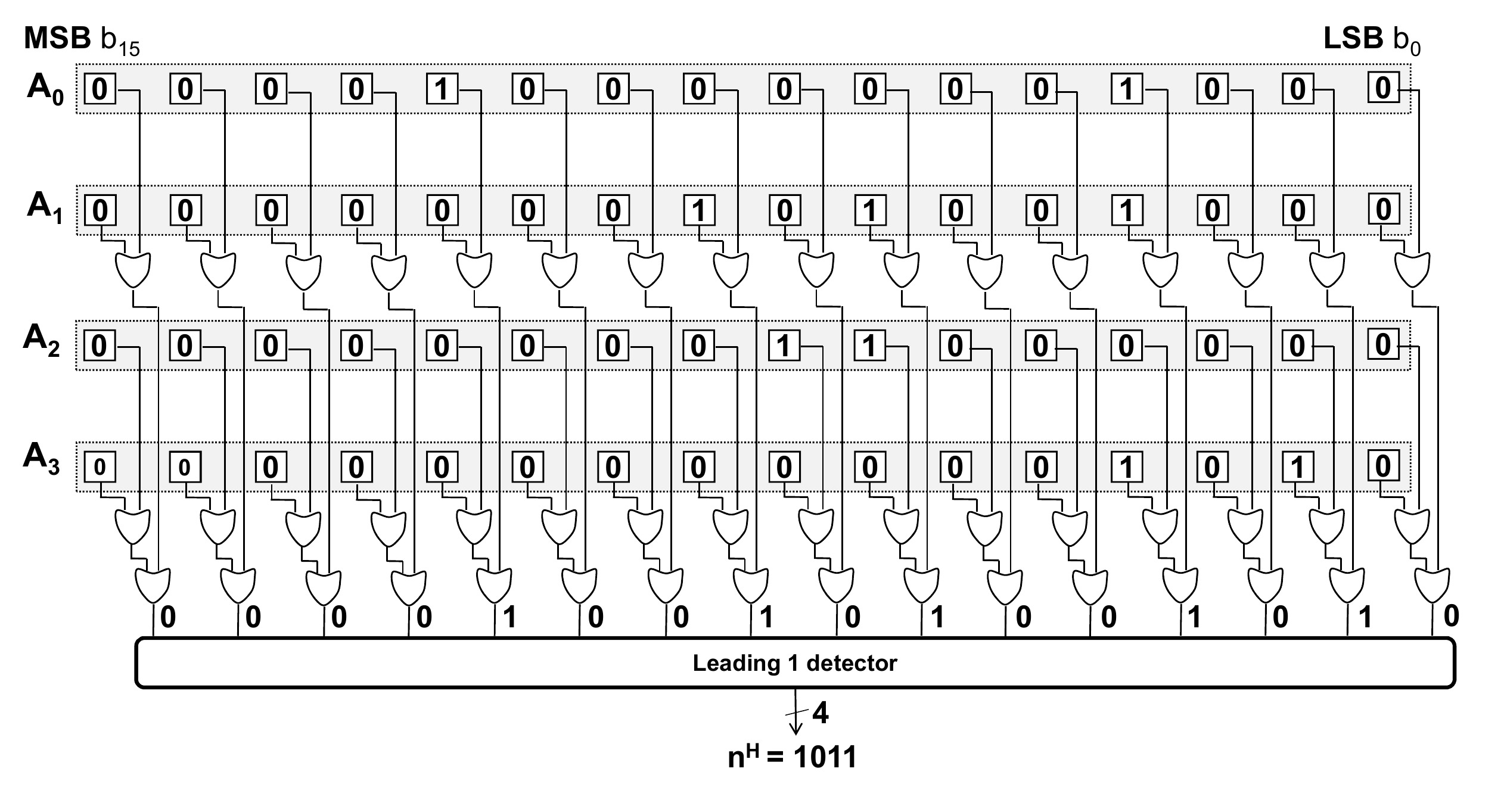}
\label{fig:pdetect}
}

\centering
\caption{Using per group precisions to reduce off-chip memory bandwidth and storage.}
\end{figure*}

\noindent\textbf{Off-chip Memory Data Container: }We encode weights and activations in groups of $N$ values. We find that $N=16$ offers a good balance between compression rate and metadata overhead. Figure~\ref{fig:memcontainer}  shows the in memory data container. For each group, we determine either statically (for weights) or dynamically (using the hardware unit of Figure~\ref{fig:pdetect}) at the output of the previous layer or at the input source for the first layer (for the activations) the precision in bits $p_i$ the group needs and store that as a prefix using 4 bits, followed by the 16 values each stored using $p_i$ bits. In addition, to avoid storing zero values altogether, we use a 16b vector with one bit per original value to identify and store only the non-zero values off-chip.  In total, this scheme requires $4+16$ bits of metadata per group of 16 values. Uncompressed these 16 values occupy 256b so the metadata overhead is far less than the benefits obtained from compressing the values for typical cases. To simplify the hardware needed, the per-group memory container can be expanded so that in total it occupies a size that is a multiple of the memory interface (64b in our experiments).

\noindent\textbf{Detecting the Per Group Precisions for Activations: }Figure~\ref{fig:pdetect} shows how  the hardware  adjusts the precision  at runtime for an example group of four 16b activations $A_0$ through $A_3$. 
It trims the unnecessary prefix bits by detecting the most significant bit position needed to represent all values within the group.  The example activations can all be represented using just 12 bits as the highest bit position $n_H$ a 1 appears is position 11. The hardware calculates 16 signals, one per bit position, each being the OR of the corresponding bit values across all four activations. The implementation uses  OR trees to generate these signals.  
 A ``leading 1'' detector identifies the most significant bit that has a 1, and reports its position in 4 bits. 
 The same method can detect  $n_L$  -- the  trailing  bit position needed to eliminate zero suffix bits. 
 In practice, adjusting $n_L$ dynamically resulted in an negligible reduction in effective precision of less than 0.4\%; profiling  effectively filters the less significant noisy bits which affect all activations regardless of  magnitude~\cite{judd:reduced}. Accordingly, we use the profile-derived $n_L$ per layer. Since our networks used the ReLU all activation values are positive. The detector can be extended to handle negative values 
by converting them first to a sign-magnitude representation, and placing the sign at the rightmost (least significant) place. This is useful for weights and for networks that use activation functions that attenuate but not remove negative values~\cite{ELU,PRELU}. 

\noindent\textbf{Memory Layout and Access Strategy: }To minimize off-chip energy, we can size the on-chip memory buffers so that each weight and activation is accessed from off-chip once per layer~\cite{kevinmemory}. In this case, the off-chip access stream is linear and contiguous and the different size of each container is no challenge. When the on-chip memory is not sufficiently large, we have to access some activations multiple times. The random access points can be easily identified at runtime when accessing them for the first time. Generally, for any given dataflow 
it will be possible to identify loop points either statically for the weights or dynamically for the activations and pre-record those in a small separate table.

\noindent\textbf{Decompression/Compression:} We use several decompression engines operating in parallel as per Figure~\ref{fig:memarch}. As data is read from off-chip in chunks of 64b, the memory controller inspects the metadata header and distributes the data packets containing the values to the decompression engines each serving a different subset of the on-chip memory banks. The decompression engines expand the values to the format used in the on-chip memories and do so serially, one value at a time, avoiding the use of wide crossbars or shuffling networks. 
At the output of each layer, the output activations are assembled in groups in each bank and are encoded accordingly using the precision detected by the hardware of Figure~\ref{fig:pdetect}. The memory controller writes the resulting data containers as they become available. 

\noindent\textbf{Reducing On-Chip Communication: } The precision reduction unit of Figure~\ref{fig:pdetect} can reduce communication when moving values on-chip as long as we use a bit-serial communication channel per value: values are trimmed just before the communication channel. A wide communication channel can be used to send multiple values concurrently.

\noindent\textbf{Reducing On-Chip Storage: } We can encode values in on-chip memory using the virtual column method of Judd et al.,~\cite{ProteusICS16}. Values within the same precision group are spread over multiple virtual columns, one value per column. A separate virtual column is needed to specify the precision for each group. Evaluating this option is left for future work.

\begin{figure*}[t]
\centering
\subfloat[][\BASE Tile]{
\centering
\includegraphics[width=0.20\textwidth]{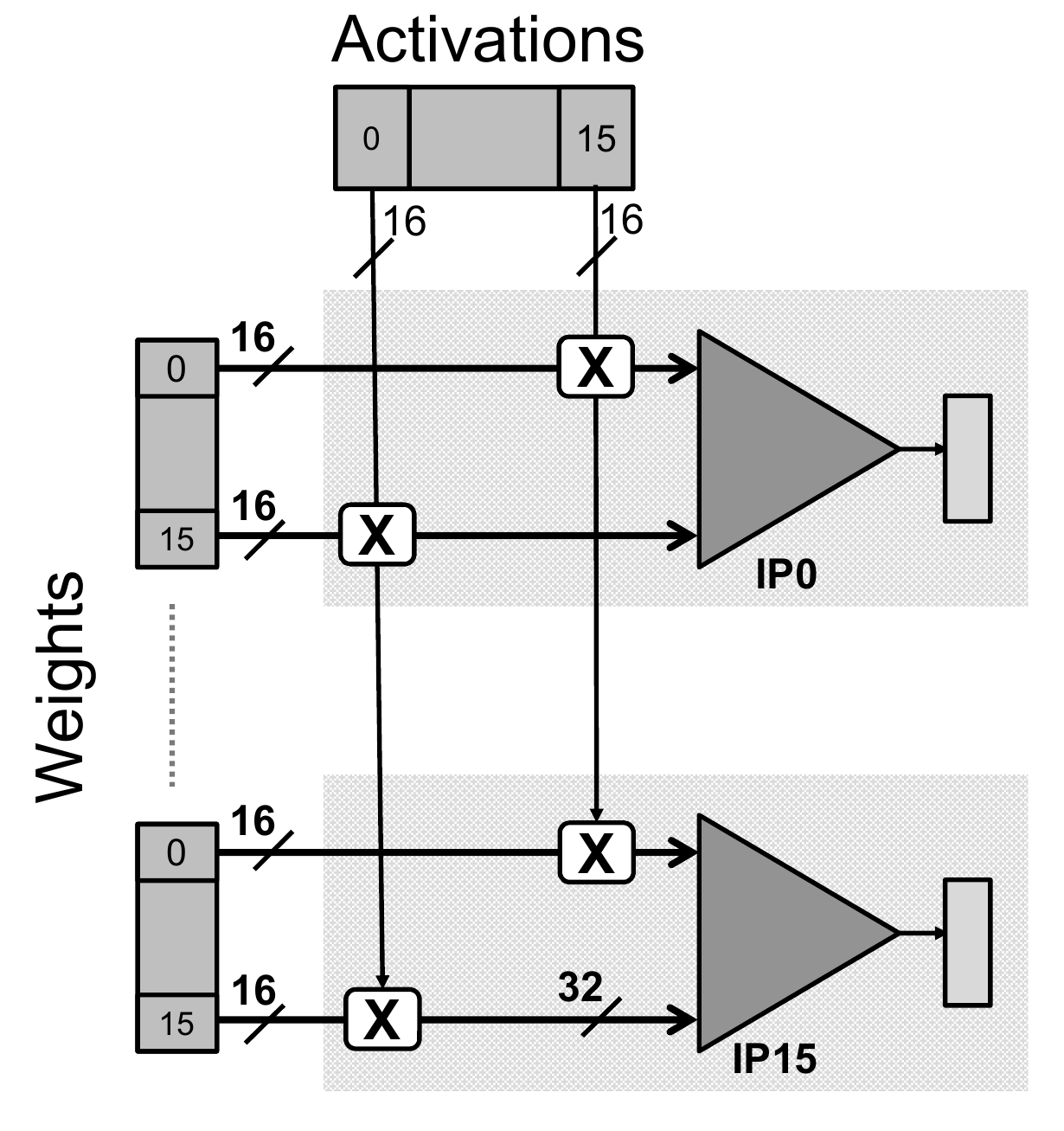}
\label{fig:basetile}
}
\subfloat[][\STRTITLE Tile]{
\centering
\includegraphics[width=0.20\textwidth]{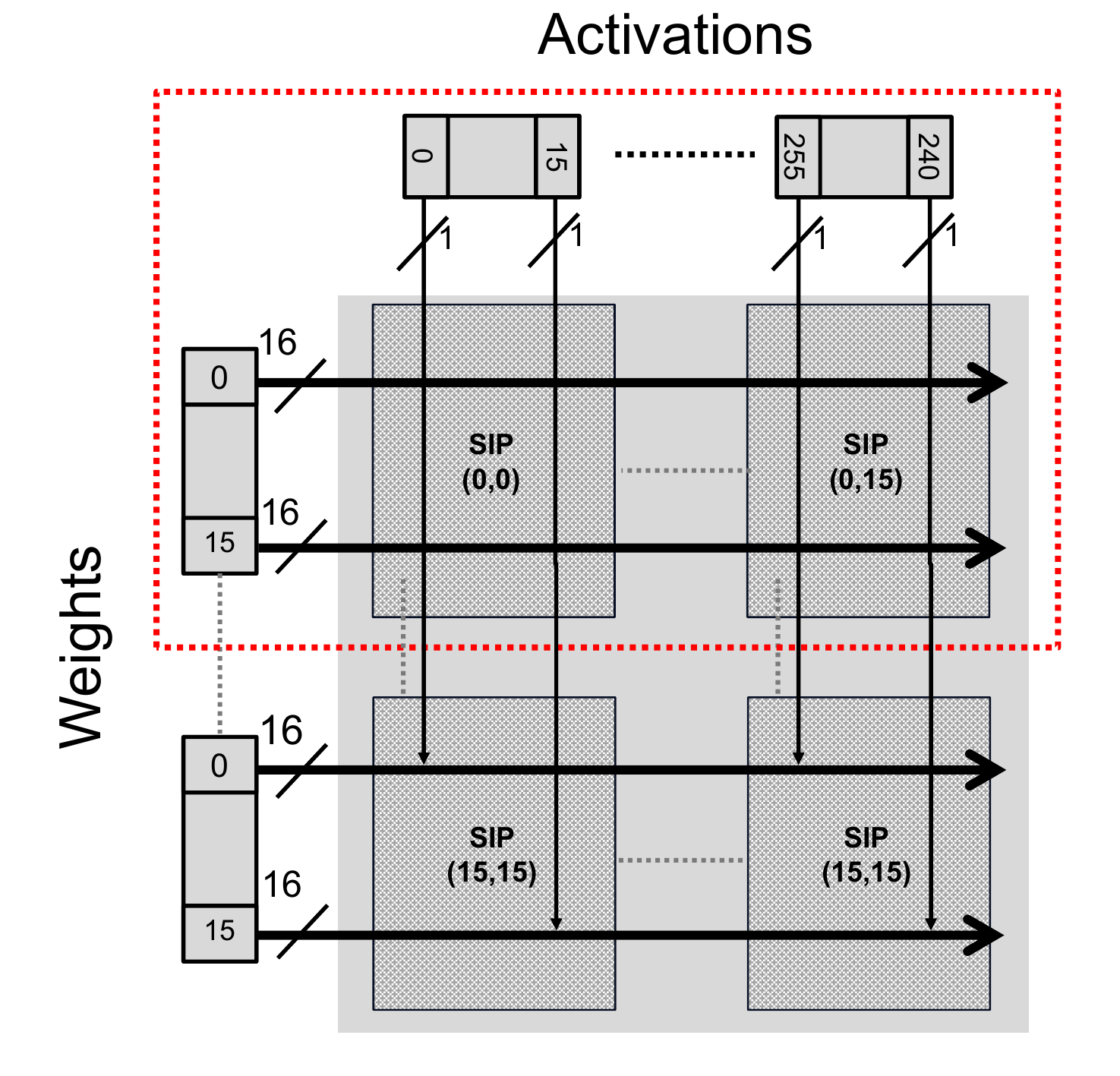}
\label{fig:str:tile}
}
\subfloat[][\STRTITLE: A Row of SIPs]{
\centering
\includegraphics[width=0.25\textwidth]{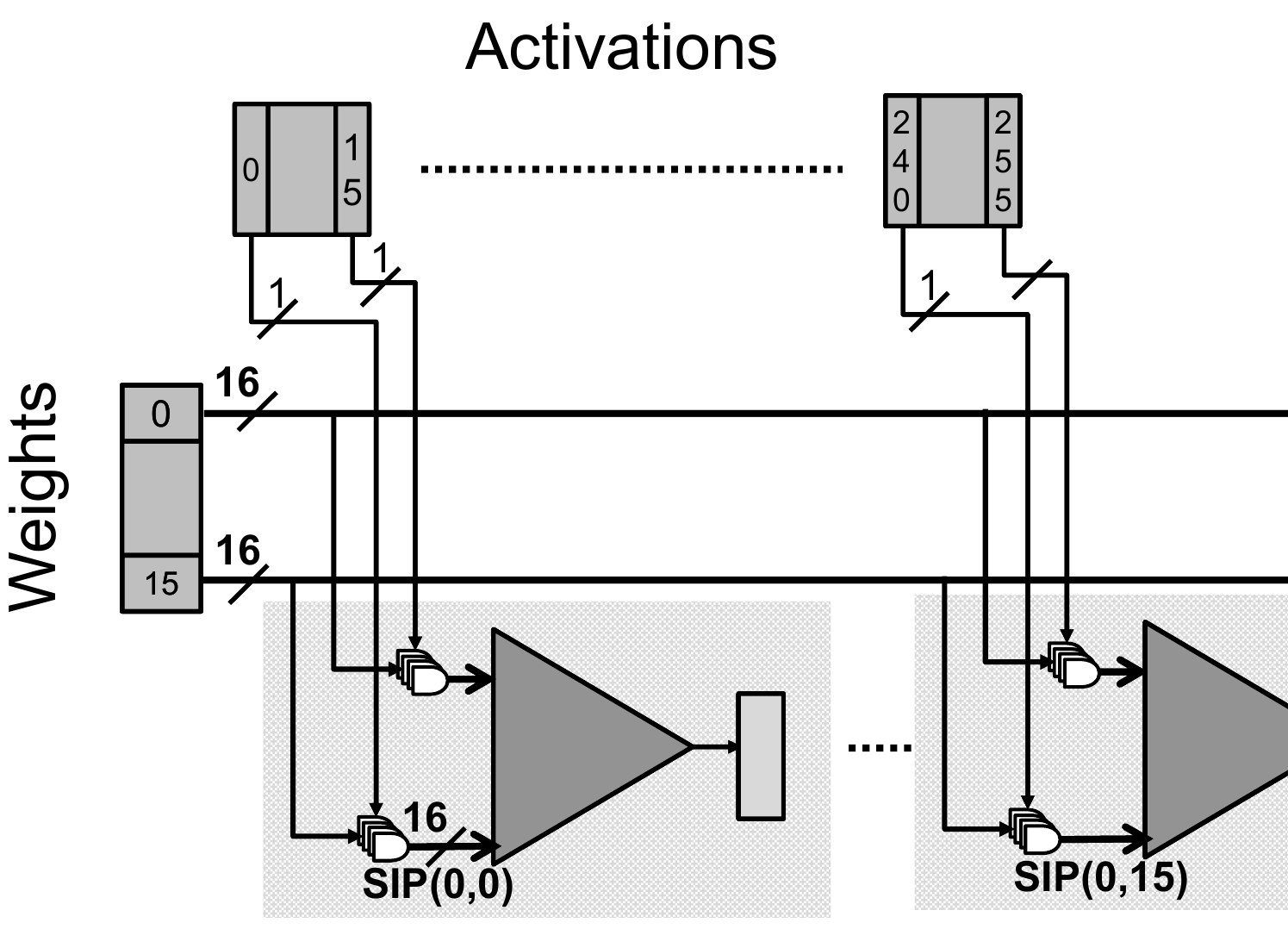}
\label{fig:str:tilerow}
}
\subfloat[][\DSTRTITLE]{
\centering
\includegraphics[width=0.13\textwidth]{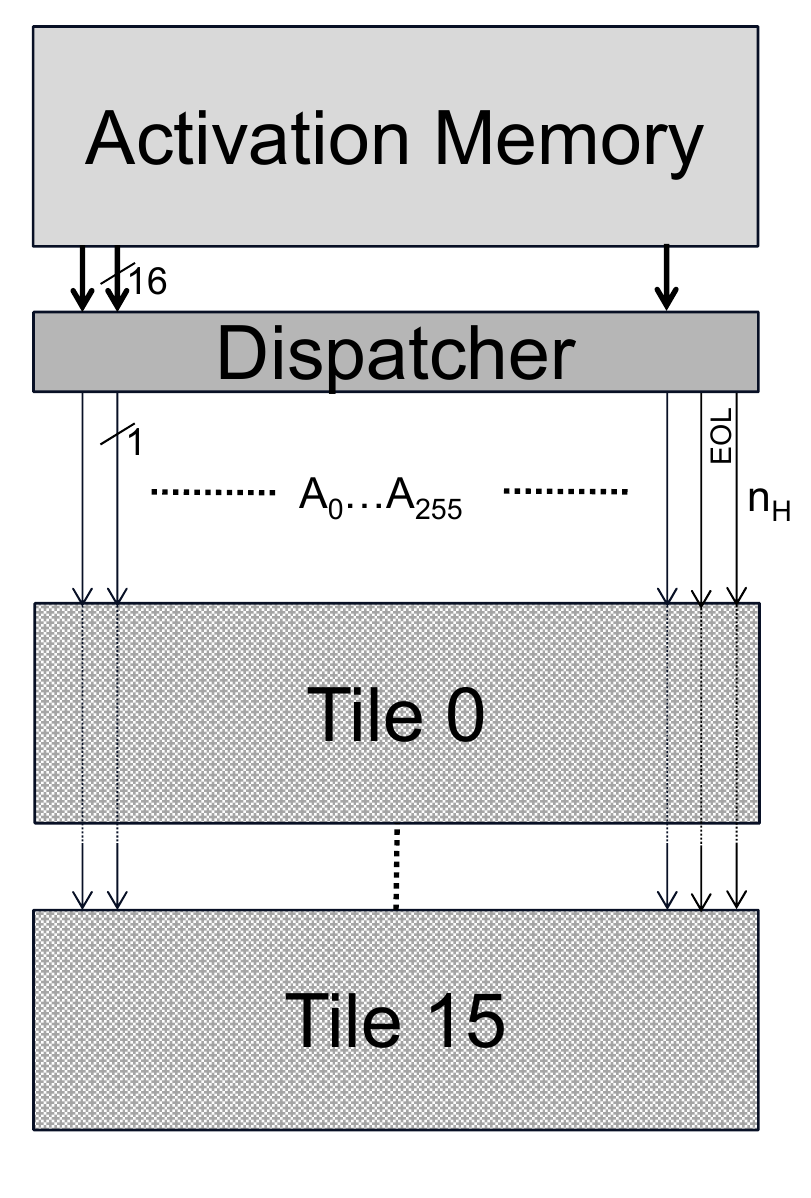}
\label{fig:dstr:overview}
}
\caption{(a) Baseline bit-parallel accelerator. (b)-(d): \DSTRTITLE: Converting \STR to dynamically reduce activation precisions.}
\label{fig:dstr}
\vspace{\floatMargin}
\end{figure*}

\section{Reducing Execution Time}
\label{sec:dstr}
The fine-grain precision requirement variability of the activations and weights can be used to also boost performance and energy efficiency. Specifically, this section presents DPred Stripes or \DSTR, an accelerator whose performance in convolutional layers scales proportionally with the inverse of the precision used 
for activations. \DSTR is unique in that it adjusts the precision on-the-fly to meet the needs of only those activations that are currently being processed. Fortunately, \DSTR can be implemented as a modest extension over the previously proposed \STRL ~\cite{Stripes-MICRO}. Simplicity and low cost are major advantages for any new hardware proposal. We first review \STR and an equivalent fixed-precision accelerator, and then explain the changes needed to enable dynamic precision reduction. 

\subsection{\STRLTITLE and Bit-Parallel Accelerator}
\label{sec:baseline}
For clarity this discussion assumes the previously described configuration of a \STR chip with 16 tiles, each processing 16 filters and 16 weights per filter and where the maximum precision is 16b. Figure~\ref{fig:basetile} shows the tile of an equivalent fixed-precision bit-parallel accelerator \BASE which  is inspired by DaDianNao~\cite{diannao}. For clarity in this discussion we assume that all data per layer fits on-chip (Google's TPU also uses large on-chip memories~\cite{TPUISCA17}).  Section~\ref{sec:evaluation}  considers the effects of reduced on-chip storage and of realistic off-chip memory systems.   A \textit{Dispatcher} fetches the activations from a central 4MB Activation Memory (AM) whereas a 2MB per tile slice of a Weight Memory (WM) provides the 4K weight bits that are needed per tile. In \BASE, the dispatcher broadcasts 16 activations, or 256 bits per cycle. Each cycle, each \BASE tile multiplies these 16 activations with 16 groups, each of 16 weights, and accumulates the results into 16 output activations. Each group of weights corresponds to a different filter and there is one weight per input activation in each group. Each \BASE tile contains 16 inner-product units ($IP_{0...15}$).
Each IP has several output accumulators.
\noindent\textbf{Data Reuse:} \BASE exploits data reuse both spatially and temporarily. Specifically each 
activation is reused across all filters, $16\times$ per tile. In addition, the output accumulators can be used used to reuse weights and activations over time by multiplexing the calculation of several output activations per tile. 

In \STR, each cycle the dispatcher broadcasts 16 sets of 16 activations  to all tiles \textit{bit-serially} for a total of 256 bits per cycle. Each activation set corresponds to a different input window so that the same 16 weights can be reused across groups to yield 16 output activations. As Figure~\ref{fig:str:tile} shows, each tile contains a grid of $16\times16$ Serial Inner-Product units (SIPs). Each SIP performs 16 $1b\times16b$ multiplications followed by a reduction. The SIPs along the same column share the same group of 16 single-bit activations, while the SIPs along the same row share the same 16 16b weights, as each SIP produces an output activation corresponding to a different window. Figure~\ref{fig:str:tilerow} shows one such row of SIPs. Overall, each tile accepts 256 input activations and $256\times16=4K$  weight bits per cycle, maintaining the same number of external wire connections as \BASE. 
Before processing a layer \STR expects software to specify the required precision, that is the positions of the most significant and of the least significant bits (MSB and LSB respectively), $n_H$ and $n_L$. \STR uses this precision for all activations within the layer.
Since \STR processes 256 activations bit-serially over $P_a$ cycles, it can ideally improve performance by $16/P_a$ over \BASE.
\noindent\textbf{Data Reuse: }\STR further boosts data reuse in space and time: 1)~each weight is reused across the 16 SIPs per row, and 2)~weights are reused over the multiple cycles it takes to calculate their product with the corresponding activation bit-serially.

\subsection{\DSTRLTITLE Architecture}
\label{DSTRArch}
The modest changes needed over \STR to implement dynamic precision reduction are:~1) introducing a mechanism for detecting the precision needed per activation group, 2)~adding a method of communicating the precision to the tiles, and 3)~modifying the SIPs to appropriately handle starting the calculation \textit{per group} at any $n_H$ bit position.

\noindent\textbf{Precision Detection: }
Figure~\ref{fig:dstr:overview} shows the \DSTR organization. A \textit{dispatcher} reads a group of 256 activations from AM, however before communicating their values bit-serially, it first inspects them detecting the precision needed. It then communicates this precision as a 4-bit offset and as a single \textit{end of group} signal.
The precision is detected using the precision detection unit of Figure~\ref{fig:pdetect}.

To process an activation group, the dispatcher sends $n_H$ as the starting offset. The tiles decrement this offset every cycle. The dispatcher signals, using an \textit{end of group} wire, the last cycle of processing for this group when the current offset becomes equal to $n_L$. 

\noindent\textbf{Granularity: }
The \DSTR configurations we study detect precision for all 256 concurrently processed activations. Other arrangements are possible. For example, precision can be detected per group of 16 activations corresponding to SIPs along the same column. 

\noindent\textbf{Modified Serial Inner-Product Unit: }
The only modification needed to the SIP is the introduction of a shifter at the output of the adder tree. This shifter adjusts the adder tree output so that it can be accumulated with the running sum aligning it at the appropriate bit position. This is necessary since starting position varies per activation group.

\newcommand{\fix}{\marginpar{FIX}}
\newcommand{\new}{\marginpar{NEW}}




\subsection{Fully-Connected Layers}
\label{sec:trt}
While \STR and \DSTR exploit precision variability for convolutional layers they do not do so for fully-connected layers. As a result, performance for fully-connected layers with \STR and \DSTR remains practically the same as that of \BASE, but energy-efficiency suffers. This section motivates further extending \STR and 
\DSTR to exploit precision variability to boost performance and energy efficiency for fully-connected layers. We motivate this change by showing that: 1)~indeed energy efficiency suffers in fully-connected layers (Section~\ref{sec:trt:ee}), and 2)~precisions vary for weights in fully-connected layers (Section~\ref{sec:trt:prec}). The aforementioned results motivate the \TRTL (\TRT) extension that improves performance and energy efficiency for fully-connected layers and which complements \STR and \DSTR. For clarity this section presents \TRT as an extension over \STR. 

\label{sec:trt:ee}

While \STR's performance in fully-connected layers is virtually identical to \BASE, its energy efficiency is on average $0.73\times$ with very little variation across networks. While in image classification workloads fully-connected layers represent less than 10\% of the overall execution time, in other applications this is not the case. Accordingly, it is desirable to improve energy efficiency for these layers as well. 

\label{sec:trt:prec}

\sloppy
The per layer precision profiles presented here were found via the methodology of Judd \textit{et al.}~\cite{judd:reduced}. For the image classification convolutional neural networks, Caffe~\cite{caffe} was used to measure how reducing the precision of each fully-connected layer affects the network's overall \textit{top-1} prediction accuracy over 5000 images. The networks are taken from the Caffe Model Zoo~\cite{model-zoo} and are used as-is without retraining.
For fully-connected layers precision exploration was limited to cases where both $P_w$ and $P_a$ are equal (the weight and activation precision ranges differ). 
 For convolutional layers only the activation precision is adjusted since none of the designs we consider can further boost performance when reducing the weight precisions. While reducing the weight precision for convolutional layers can reduce their memory footprint~\cite{ProteusICS16}, we do not explore this option in this evaluation.
For NeuralTalk, we measure BLEU scores when compared with the ground truth. For Denoise, we measure PSNR, and define >99\% accuracy as a drop of no more than 0.04dB. 

Table~\ref{tab:TRN_precisions} reports the resulting per layer precisions. The ideal speedup columns report the performance improvement that would be possible if execution time could be reduced proportionally with precision compared to a 16-bit bit-parallel baseline. For the fully-connected layers, the precisions required range from 8 to 10 bits and the potential for performance improvement is $1.64\times$ on average. 
Given that the precision variability for fully-connected layers is relatively low (ranges from 8 to 11 bits) one may be tempted to conclude that an 11-bit \BASE variant may be an appropriate compromise. However, given that the precision variability is much larger for the convolutional layers (range is 5 to 13 bits) the performance with a fixed precision datapath would be far below the ideal. 
Section~\ref{sec:evaluation} shows that the incremental cost  of \TRT over \STR is well justified given the benefits.

\begin{table}
\scriptsize
\centering
    \caption{
Per layer precision profiles needed to maintain the same accuracy as in the baseline. \textit{Ideal}: Potential speedup with bit-serial processing of activations over a 16-bit bit-parallel baseline without dynamic adaptation.
}
\label{tab:TRN_precisions}

    \scriptsize
    \begin{tabular}{|l|p{4cm}|r|}
\hline
\multicolumn{3}{|c|}{\textbf{Convolutional Layers}} \\ 
\hline
 \textbf{Network}                                     & \textbf{Activation Precision in Bits}                                 & {\textbf{Ideal} }  \\ 
 \hline
&{ \textbf{100\% Accuracy} }                                                                               &   \textbf{Speedup}                                      \\ 
\hline
{AlexNet}                         & 9-8-5-5-7                                                             & {2.38}               \\ 
\hline
{VGG\_S}                           & 7-8-9-7-9                                                             & {2.04}               \\ 
\hline
{VGG\_M}                           & 7-7-7-8-7                                                             & {2.23}               \\ 
\hline
{VGG\_19}                          & 12-12-12-11-12-10-11-11-13-12-13-13-13-13-13-13                       & {1.35}               \\ 
\hline
{NeuralTalk}                      & -                                                                     & {-}                  \\ 
\hline
{Denoise}                         & -                                                                     & {-}                  \\ 
\hline

\multicolumn{3}{|c|}{\textbf{Fully Connected Layers } }                                                                                                                   \\\hline
                                                      & {\textbf{Act. \& Weight Precision in Bits}~~ ~}             & \textbf{Ideal}                        \\
\hline
\multicolumn{2}{|c|}{\textbf{100\% Accuracy} }                                                                                  &  \textbf{Speedup}                                       \\
\hline
AlexNet                                               & {{10-9-9}}              & {1.66}      \\
\hline
VGG\_S                                                 & {{10-9-9}}              & {1.64}      \\
\hline
VGG\_M                                                 & {{10-9-9}}              & {1.64}      \\
\hline
{VGG\_19}                  & {{10-9-9}}              & {1.63}      \\
\hline
NeuralTalk                                            & {{11 (all iterations)}} & {1.45}      \\
\hline
Denoise                                               & {{12 (all layers)}}     & {1.33}      \\
\hline
\end{tabular}
\vspace{\floatMargin}

\end{table}

\subsubsection{\TRTLTITLE} 
\label{sec:tartan}

\begin{figure}
\centering
\subfloat[][A row of a \TRTTITLE tile]{
\includegraphics[width=0.35\textwidth]{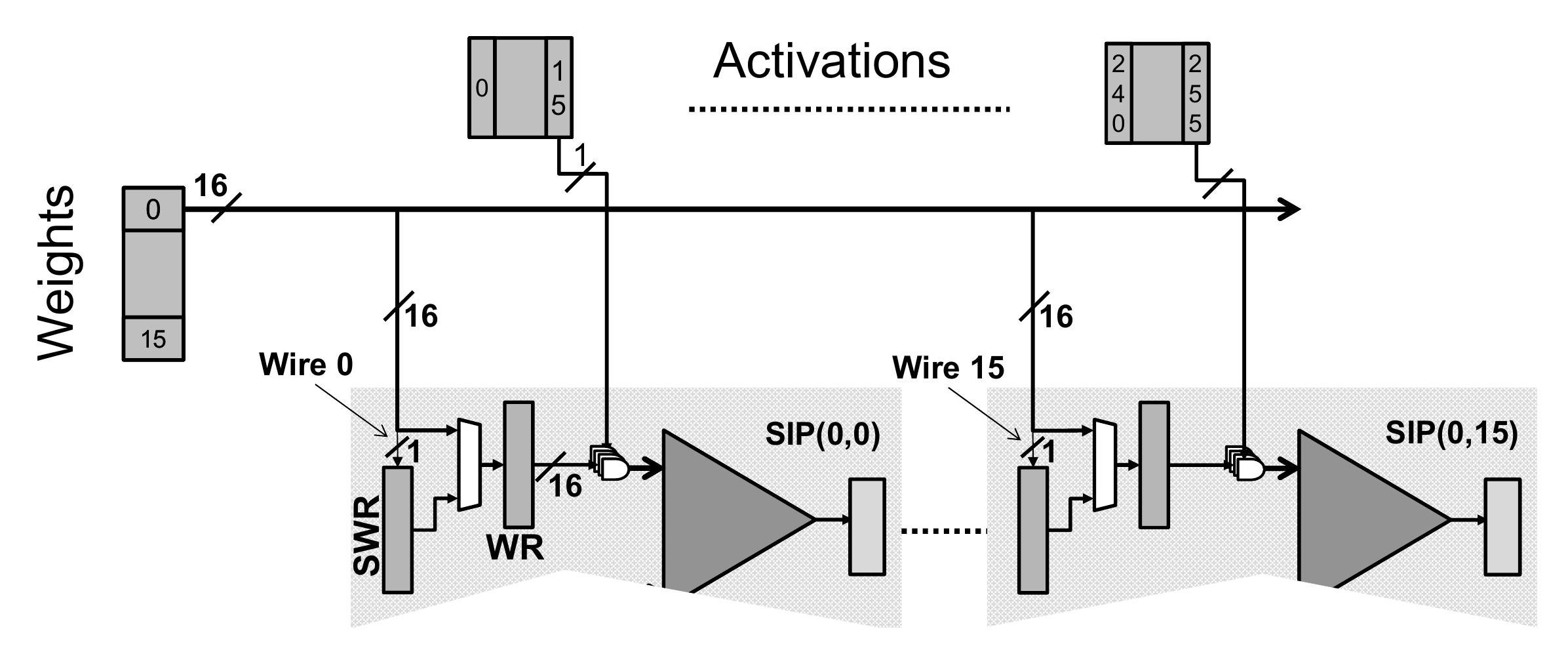}
\label{fig:trt:concept}
}\\
\subfloat[][\TRT's SIP.]{
\includegraphics[scale=0.35]{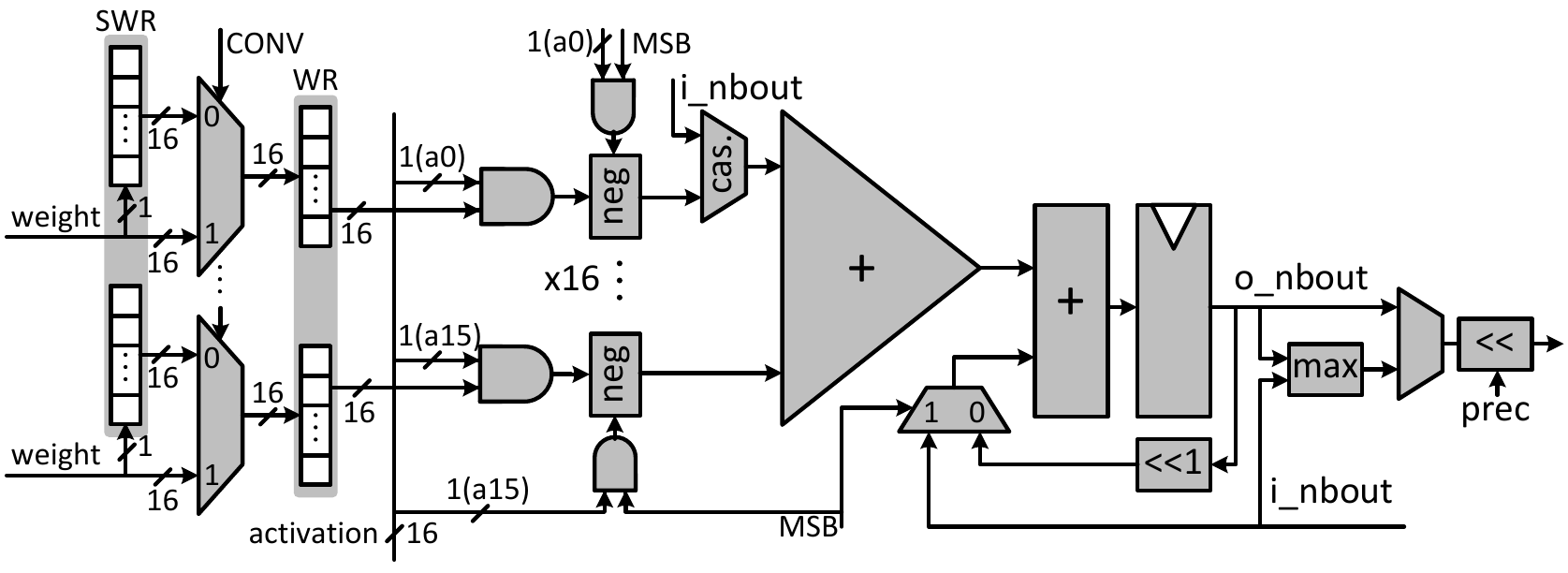}
\label{fig:trt-sip}
}
\caption{\TRT Tile and SIP.}
\end{figure}

Unfortunately, since there is no weight reuse in fully-connected \DSTR cannot reuse the same weight over multiple windows. In \DSTR  performance in fuly-connected layers is limited by the number of cycles needed to load a different set of weights per SIP column~\cite{Stripes-MICRO}. \TRT overcomes this limitation and boosts performance even for fully-connected layers by exploiting weight and activation precisions. 
We use Figure~\ref{fig:trt:concept} to explain the concept behind \TRT's operation. The figure shows a row of SIPs (similar to Figure~\ref{fig:str:tilerow}) and focuses on the first weight input only. Whereas in \STR the 16b weight input to the AND gates was coming directly from the WM connection here it is connected to a 16b \textit{Weight Register} (WR). A multiplexer selects from where the WR can load its contents. The first option is to do so directly from the WM wires as in \STR. This maintains the functionality needed for convolutions. An extra pipeline stage is needed to accommodate loading first to WR (stage 1) and then ``multiplying'' with the activation (stage 2). The second option is to load WR from the 16b \textit{Serial Weight Register} (SWR). The SWR has a single input wire connection to the Weight Memory which is different per SIP. The first SIP in the row connects to wire 0 while the last to wire 15. SWR is a serial-load register and it can load a new weight value of $p$ bits bit-serially over $p$ cycles. Given that each SWR connects to a different Weight Memory wire, they can all concurrently load a different $p$ bit weight value over the same $p$ cycles. Each SIP can then copy its SWR held weight into its WR and proceed with processing the corresponding activations bit serially. Concurrently with processing the activations, the SWR can proceed to load the next set of weights. Accordingly, loading weights into the SWRs and processing the ones in the WRs form a two ``stage'' pipeline. The first ``stage'' of this pipeline requires $P_w$ cycles to load a new set of weights since the weights are loaded bit-serially. The second ``stage'' of the pipeline requires $P_a$ to process the current set of weights and activations since it multiplies activations bit-serially. As a result, it is the maximum precision of the weights or the activations that will dictate the number of cycles needed to process each group of activations.

Since there are 16 weights per SIP, the SWR and WR are implemented each as a vector of 16 16-bit subregisters. The remainder of this section explains how \TRT processes convolutional and fully-connected layers. For clarity, in what follows the term \textit{brick} refers to a set of 16 elements of a 3D activation or weight array input which are contiguous along the \textit{channel} dimension, e.g., $a(x,y,i)...a(x,y,i+15)$. Bricks will be denoted by their origin element with a $B$ subscript, e.g., $a_B(x,y,i)$.  The size of a brick is a design parameter.

\noindent\textbf{Convolutional Layers:} Processing is identical to \STR and starts by reading in parallel 256 weights from the Weight Memory as in \STR, and loading the 16 per SIP row weights in parallel to all SWRs in the row.

\noindent\textbf{Fully-Connected Layers:} Processing starts by loading bit-serially and in parallel over $P_{w}$ cycles 4K weights into the 256 SWRs, 16 per SIP. Each SWR per row gets a different set of 16 weights as each subregister is connected to one out of the 256 wires of the Weight Memory output bus for the SIP row (as in \BASE\ there are  $256\times 16=4K$ wires).
Once the weights have been loaded, each SIP copies its SWR to its SW and multiplication with the input activations can then proceed bit-serially over $P_{a}^L$ cycles. Assuming that there are enough output activations so that a different output activation can be assigned to each SIP, the same input activation brick can be broadcast to all SIP columns. That is, each \TRT tile processes one activation brick $a_B(i)$ bit-serially to produce 16 output activation bricks $o_B(i)$ through $o_B(i\times 16)$, one per SIP column. Loading the next set of weights can be done in parallel with processing the current set, thus execution time is constrained by $P_{max} = max(P_{a},P_{w})$. Thus, a \TRT tile produces 256 partial output activations every $P_{max}$ cycles, a speedup of $16/P_{max}$ over \BASE since a \BASE tile always needs 16 cycles to do the same.

\noindent\textbf{Cascade Mode:} \TRT is underutilized in fully-connected layer with less than 4K output activations. Some networks have layers with as little as 2K outputs. To avoid underutilization, the SIPs along each row are cascaded into a daisy-chain, where the output of one can feed into an input of the next via a multiplexer. This way, the computation of an output  can be sliced over the SIPs along the same row: Each SIP processes only a portion of the input activations resulting into several partial output activations along the SIPs on the same row. Over the next $np$ cycles, where $np$ the number of slices used, the $np$ partial outputs are added. The user can chose any number of slices up to 16, so that \TRT can be fully utilized even when there are just 256 outputs. 

\noindent\textbf{Other Layers:} \TRT like \STR can process the additional layers needed by the studied networks. The tile includes hardware support for max pooling similar to \STR. An activation function unit is present at the output in order to apply nonlinear activations before writing back to AM.

\noindent\textbf{SIP and Other Components: }
\label{sec:computeunit}
Figure \ref{fig:trt-sip} shows \TRT's SIP which multiplies 16 activation bits, one bit per activation, by 16 weights to produce an output activation. Two registers, a Serial Weight Register (SWR) and a Weight Register (WR), each contain 16 16-bit sub-registers. Each SWR sub-register is a shift register with a single bit connection to one of the weight bus wires that is used to read weights bit-serially for FCLs. Each WR sub-register can be parallel loaded from either the weight bus or the corresponding SWR sub-register to process convolutional or fully-connected layers respectively. The SIP includes 256 2-input AND gates that multiply the weights in the WR with the incoming activation bits, and a $16\times 16b$ adder tree that sums the partial products. A final adder plus a shifter accumulate the adder tree results into the output register.
A multiplexer at the first input of the adder tree implements the cascade mode supporting slicing the output activation computation along the SIPs of a single row.  Each SIP also includes a comparator (max) to support max pooling layers. A shifter between the output of the adder tree and the input to the accumulator can be added to support \DPR.

As in \STR there is a central AM and 16 tiles. A \textit{Dispatcher} unit is tasked with reading input activations from AM always performing eDRAM-friendly wide accesses. It transposes each activation and communicates each a bit a time over the global interconnect. For convolutional layers the dispatcher has to maintain a pool of multiple activation bricks, each from different window, which may require fetching multiple rows from AM. However, since a new set of windows is only needed every $P_{a}$ cycles, the dispatcher can keep up for the layers studied. For fully-connected layers one activation brick is sufficient. A \textit{Reducer} per title is tasked with collecting the output activations and writing them to AM. Since output activations take multiple cycles to produce, there is sufficient bandwidth to sustain all 16 tiles.

\noindent\textbf{Coarser Bit Granularity Processing: }
\label{sec:twobit}
To improve \TRT's area and power efficiency, the number of activation bits processed at once can be adjusted at design time. Such designs need fewer SIPs and shorter wires. %
 However, they forgo some of the performance potential as they force the activation precisions to be a multiple of the number of bits that they process per cycle. 

\subsection{\DPRTITLE Loom}
Dynamic precision variability can also benefit Loom~\cite{Sharify:2018:LEW:3195970.3196072}, an architecture that exploits precision variability both in activations and weights which is faster and more energy efficient than \STR for smaller configurations. 
The original design, which uses per-layer precisions, can be extended utilizing the mechanism described in Section~\ref{DSTRArch} to take advantage of per group precisions for both weights and activations.

\section{Evaluation}
\label{sec:evaluation}

We first assume that all data fits on chip using the memory configuration of DaDianNao~\cite{DaDiannao}. Having identified a reasonable compute configuration we then study restricted on-chip memory hierarchies and the effect of off-chip memory demonstrating that \DPR greatly helps with alleviating stalls due to off-chip accesses.

All accelerators were modeled using the same methodology for consistency. A custom cycle-accurate simulator models execution time. Computation was scheduled as described in \STR to maximize energy efficiency for \BASE~\cite{Stripes-MICRO}. 
To estimate power and area, all designs were synthesized with the Synopsys Design Compiler~\cite{synopsysDC} for a TSMC 65nm library and laid out with Cadence Encounter. 
Circuit activity was captured with ModelSim and fed into Encounter for power estimation. 
All designs operate at 980 MHz. The SRAM activation buffers were modeled using CACTI~\cite{Muralimanoharcacti6.0:}. The AM and WM eDRAM area and energy were modeled with \textit{Destiny}~\cite{destiny}. Three design corner libraries were considered prior to layout. The typical case library was chosen for layout since bit-serial designs are affected less by the worst-case design corner. Accordingly, the relative benefits with \DPR over \BASE are underestimated. \DSTR and \TRT improve energy efficiency and performance for all design corners.

Given that \DSTR improves performance only for conv. layers, and that \TRT performs identically to \DSTR for  conv layers while improving performance only for fully-connected layers, a different set of DNNs is used during the evaluation of the two techniques. 
For \DSTR measurements were performed over a set of ImageNet Classification CNNs: AlexNet, GoogleNet, Nin, VGG\_S, VGG\_M, and VGG\_19. For these networks conv. layers account for more than 90\% of the execution time. For \TRT the evaluation omits NiN and GoogleNet since the former has no fully-connected layers and their usage in the latter is not significant. Instead NeuralTalk~\cite{DBLP:journals/corr/KarpathyF14} and Denoise~\cite{plain_NN_denois} are included that are dominated by fully-connected layers. NeuralTalk uses long short-term memory to automatically generate image captions. Denoise is a 5-layer Multilayer Perceptron that implements image denoising aiming to reproduce the results of the state-of-the-art BM3D denoising algorithm~\cite{BM3D}.

\subsection{\DSTRTITLE} 
\label{sec:eval:dstr}

\begin{figure}
\centering
\subfloat[][Speedup with a group of 256]{
\includegraphics[width=0.4\textwidth]{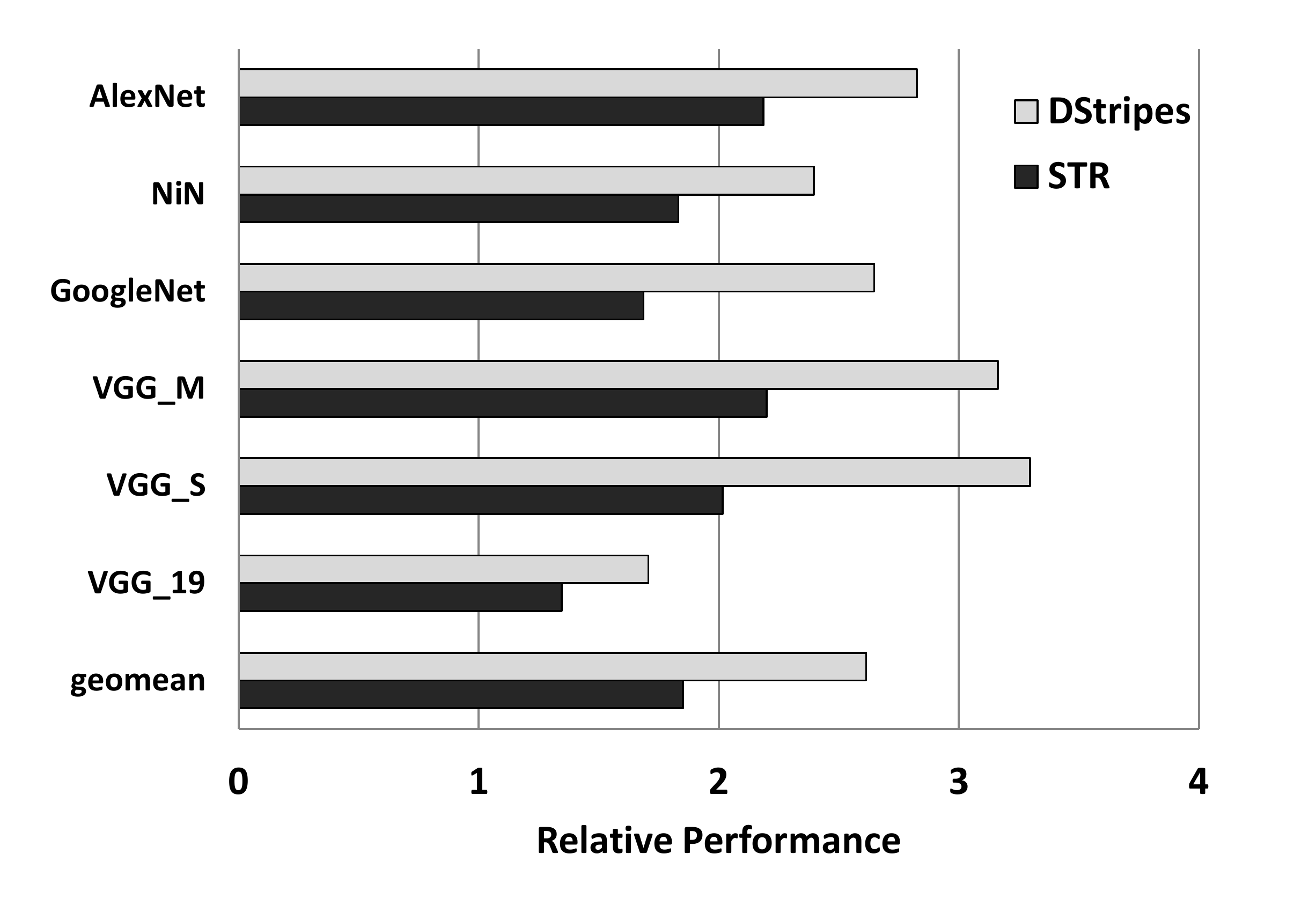}
\label{fig:eval:dstr:base}
}\\
\subfloat[][\DSTRTITLE and SCNN in pruned networks.]{
\includegraphics[width=0.4\textwidth]{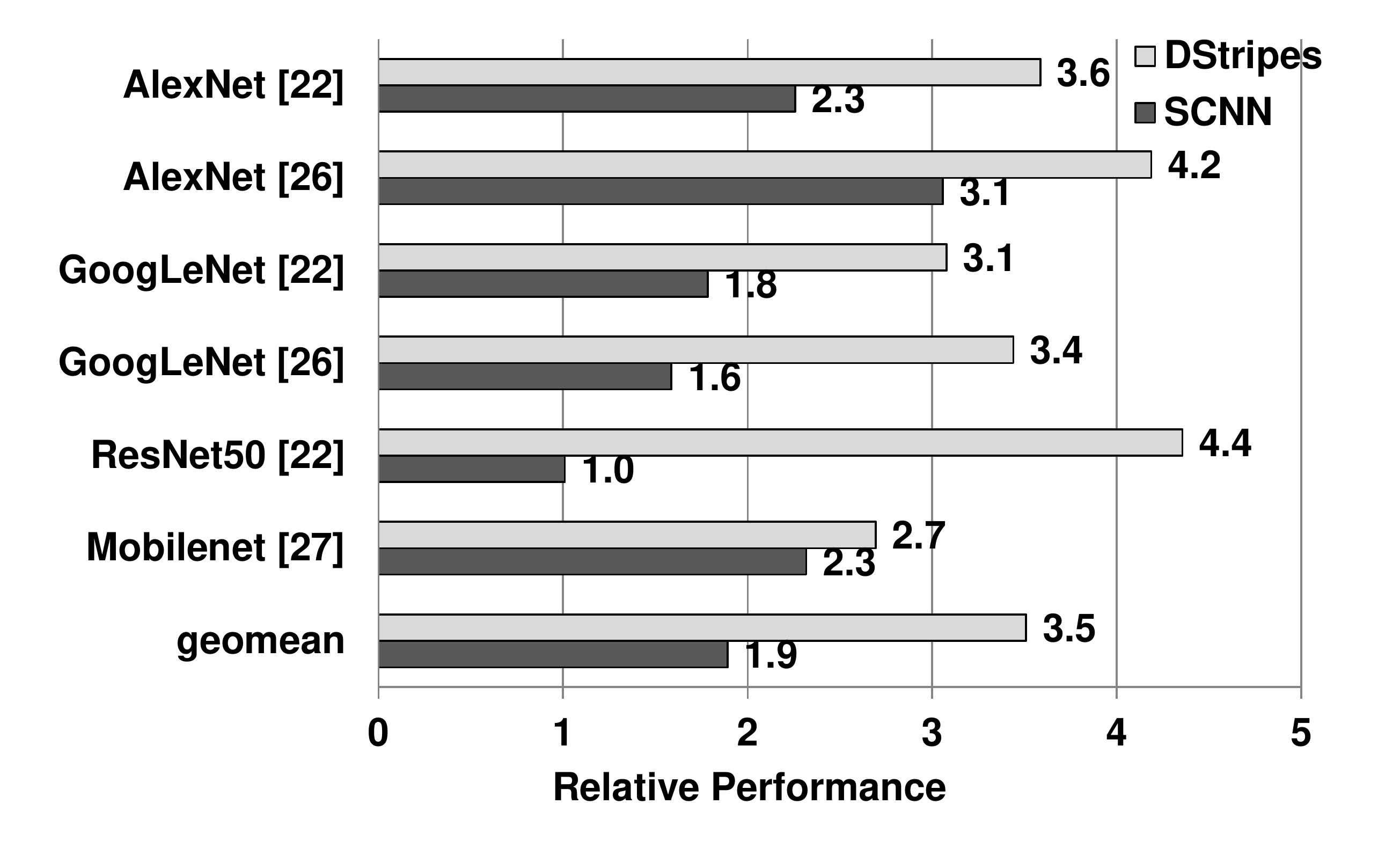}
\label{fig:eval:pruned}
}
\caption{\DSTRTITLE: Speedup over \BASE in Convolutional Layers}
\vspace{\floatMargin}
\end{figure}

 \noindent\textbf{Performance: }
Figure~\ref{fig:eval:dstr:base} shows the resulting speedup with \DSTR alone and over the equivalently configured \BASE and \STR. Since \DSTR improves performance only for conv. layers these measurements are restricted to conv. layers only. On average, dynamic precision reduction boosts performance over \STR by 41\% since \DSTR proves $2.61\times$ faster than \BASE.  

\noindent\textbf{Energy Efficiency: }
As Table~\ref{tbl:e:dstr:ee} shows \DSTR greatly improves energy efficiency over \BASE and \STR. The dispatcher has to communicate less bits and the units have to perform fewer calculations. On average, compared to \BASE, \STR and \DSTR are $1.84\times$ and $1.38\times$ more energy efficient respectively. 

\noindent\textbf{Area: }
In the interest of space, we omit the detailed area results here and present those for the combination of \DSTR and \TRT instead in the respective section.
The overall area overhead of dynamic precision reduction is less than 1\% over \STR and as a result \DSTR is $1.32\times$ larger than \BASE. Given that \DSTR is $2.61\times$ faster, the performance over area ratio is superlinear with \DSTR. Utilizing 32\% more area in \BASE would at best increase performance proportionally. However, in practice the speedup will be a lot less due to filter and input dimensions and filter count.

\begin{table}
\centering
\scriptsize
\caption{Convolutional Layers: Energy Efficiency of \DSTRTITLE vs. \BASE and vs. \STR (higher is better)}
\label{tbl:e:dstr:ee}
\begin{tabular}{|l|l|l|}
\hline
\textbf{Network}   & \textbf{vs. \BASE} & \textbf{vs. \STR} \\ \hline
\textbf{AlexNet}   & 1.98               & 1.26              \\ \hline
\textbf{NiN}       & 1.68               & 1.27              \\ \hline
\textbf{GoogleNet} & 1.86               & 1.53              \\ \hline
\textbf{VGG\_M}    & 2.22               & 1.40              \\ \hline
\textbf{VGG\_S}    & 2.31               & 1.58              \\ \hline
\textbf{VGG\_19}   & 1.20               & 1.24              \\ \hline
\textbf{GEOMEAN}   & \textbf{1.84}      & \textbf{1.38}     \\ \hline
\end{tabular}
\vspace{\floatMargin}
\end{table}

\begin{table}[!t]
\centering
\scriptsize
\caption{
Execution time and energy efficiency improvement with \TRTEVAL compared to \BASE. 
}
\label{tab:TTT}
\begin{tabular}{|c|c|c|c|c|}
\hline
                        & \multicolumn{2}{c|}{\textbf{Fully Connected Layers}} & \multicolumn{2}{c|}{\textbf{Convolutional Layers}} \\ \hline
                        & \textbf{Perf}           & \textbf{Eff}           & \textbf{Perf}     & \textbf{Eff}      \\ \hline
AlexNet                 & 1.61                      & 1.10                     & 2.81                & 1.28                \\ \hline
VGG\_S                  & 1.61                      & 1.09                     & 3.26                & 1.49                \\ \hline
VGG\_M                  & 1.61                      & 1.10                     & 3.15                & 1.44                \\ \hline
VGG\_19                 & 1.60                      & 1.09                     & 1.70                & 0.77                \\ \hline
NeuralTalk              & 1.42                      & 1.01                     & -                   & -                   \\ \hline
Denoise                 & 1.30                      & 0.92                     & -                   & -                   \\ \hline
\textbf{geomean}      & \textbf{1.52}           & \textbf{1.05}          & \textbf{2.65}     & \textbf{1.21}     \\ \hline
\end{tabular}
\vspace{\floatMargin}
\end{table}

\subsection{Pruned Models}
{Pruning is sometimes possible and it converts several weights into zeros. Past work has exploited pruning to reduce work. Figure}~\ref{fig:eval:pruned} {compares the performance of \DSTR \ and SCNN}~\cite{SCNN} { for a set of pruned models}~\cite{SkimCaffePaper,cvpr_2017_yang_energy}{. For both designs we assume infinite off-chip bandwidth giving an advantage to SCNN since \DPR  reduces traffic more than zero compression. SCNN removes all products where the activation or the weight is zero. The reasons why \DSTR outperforms SCNN are: 1) SCNN's performance is limited by inter- and intra-tile fragmentation~\cite{SCNN}, while 2) \DSTR\ targets \textit{all} activations where the potential for work reduction is higher than the work reduction potential from targeting those values that are zero. However, Dynamic Precision Reduction is compatible with SCNN and we have experimented with replacing the bit-parallel multiply-accumulate units in SCNN with bit-serial ones while increasing the number of tiles. Unfortunately, the benefits are limited since the higher tile count exacerbates inter-tile load imbalance. Addressing this imbalance is left for future work. 
}

\subsection{\DSTRTITLE and \TRTLTITLE}
\noindent\textbf{Performance: }
We evaluate the combination of \DSTR and \TRT denoted as \TRTEVAL.
Table~\ref{tab:TTT} reports \TRTEVAL's performance and energy efficiency relative to \BASE for the precision profiles in Table~\ref{tab:TRN_precisions} separately for fully-connected layers, convolutional layers, and the whole network. Denoise has no convolutional layers. NeuralTalk uses a modified convolutional neural network (based on VGG\_16 in this implementation) to recognize objects before the LSTM stage; we do not evaluate that phase as it is similar to the other image classifiers. 

On fully-connected layers \TRTEVAL yields, on average, a speedup of {\meanspeedupFC}. 
For conv. layers \TRTEVAL improves performance by {\meanspeedupCV}. %
There are two main reasons \TRTEVAL does not reach the ideal speedup: dispatch overhead and under-utilization. During the initial $P_w$ cycles the serial weight loading process prevents any useful products to be performed. This represents less than 2\% for any given network, although it can be as high as 6\% for the smallest layers. Underutilization can happen when the number of outputs is not a power of two or lower than 256.

\noindent\textbf{Energy Efficiency}
\label{sec:eval:ee}
As Table~\ref{tab:TTT} reports, the average efficiency improvement with \TRTEVAL across all networks and layers is {\meanefficiencyALL}. \TRTEVAL is more efficient than \BASE for all layers.
Overall, efficiency primarily comes from the reduction in effective computation due to reduced precisions. Furthermore, the amount of data that has to be transmitted from the WM and the traffic between the AM and the SIPs is decreased proportionally with the chosen precision.

\noindent\textbf{Area: } 
\label{sec:eval:area}
Table~\ref{table:TRN_area} reports the area breakdown of \TRTEVAL and \BASE. Over the full chip, \TRTEVAL needs $1.51\times$ the area compared to \BASE while delivering on average a \meanspeedupALL speedup. Generally, performance would scale sublinearly with area for \BASE due to underutilization. 

\noindent\textbf{Sensitivity to Precision Resolution: }
\label{sec:eval:2b}
Table~\ref{tab:TBT} reports performance for the \TRTEVAL variant of Section~\ref{sec:twobit} that processes 2 bits per cycle in as half as many total SIPs. The previously quoted precisions are rounded up to the next multiple of two.  This design always improves performance compared to \BASE. Compared to the 1-bit \TRTEVAL performance is slightly lower, however this can be a good trade-off given the reduced cost and improved energy efficiency. Overall, there are two forces at work: There is performance potential lost due to rounding all precisions to an even number, and there is performance benefit by requiring less parallelism. The time needed to serially load the first bundle of weights is also reduced. 
The layout of \TRTEVAL's 2-bit variant requires only 26.0\% more area than \BASE (Table~\ref{table:TRN_area}) while improving energy efficiency in fully-connected layers by $1.24\times$ on average ($1.44\times$ across all layer types).

\begin{table}[t]
\centering
\scriptsize
\caption{Area Breakdown for \TRTEVAL and \BASE}
\label{table:TRN_area}
\begin{tabular}{|c|c|c|c|}
\hline
 & \textbf{\TRTEVAL} &  \textbf{\TRTEVAL 2-bit} & \textbf{\BASE} \\
\hline
\hline 
\textbf{Inner-Product Units}  & 58.60   (48.21\%)       & 38.32   (37.84\%)       & 17.85   (22.20\%) \\
\textbf{Weight Memory}        & 48.11   (39.58\%)       & 48.11   (47.50\%)       & 48.11   (59.83\%) \\
\textbf{Act. Input Buffer}    & 3.66    (3.01\%)        & 3.66     (3.61\%)       & 3.66    (4.55\%) \\
\textbf{Act. Output Buffer}   & 3.66    (3.01\%)        & 3.66     (3.61\%)       & 3.66    (4.55\%) \\
\textbf{Activation Memory}    & 7.13    (5.87\%)        & 7.13     (7.04\%)       & 7.13    (8.87\%) \\
\textbf{Dispatcher}           & 0.39    (0.32\%)        & 0.39     (0.39\%)       & -                \\\hline
\textbf{Total}                & 121.55 (100\%)          & 101.28 (100\%)          & 80.41 (100\%)    \\ \hline
\textbf{Normalized Total}     & $1.51\times$            & $1.26\times$            & $1.00\times$     \\ 
\hline
\end{tabular}
\vspace{\floatMargin}
\end{table}

\begin{table}[!t]
\centering\scriptsize
\caption{
Relative performance of the 2-bit \TRTEVAL  
}
\label{tab:TBT}
\begin{tabular}{|c|c|c|c|c|}
\hline

 \   & \multicolumn{2}{|c|}{\textbf{\vtop{\hbox{\strut Fully Connected}\hbox{\strut Layers}} }} & \multicolumn{2}{|c|}{\textbf{\vtop{\hbox{\strut Convolutional}\hbox{\strut Layers}} }}   \\
          \cline{1-5}
   &  \textbf{  \vtop{\hbox{\strut vs.}\hbox{\strut BASE}}} &\textbf{\vtop{\hbox{\strut vs.}\hbox{\strut 1b \TRTEVAL}}}  &  \textbf{\vtop{\hbox{\strut vs.}\hbox{\strut BASE}}} &\textbf{\vtop{\hbox{\strut vs.}\hbox{\strut 1b \TRTEVAL}}} \\ 
          \cline{1-5}

        AlexNet      & +58\% &  -2.06\% & +164\% &    -7.11\% \\ \hline
        VGG\_S       & +59\% &  -1.25\% & +191\%  &  -13.27\% \\ \hline
        VGG\_M       & +63\% &  +1.12\% & +181\%  &  -12.57\% \\ \hline
        VGG\_19      & +59\% &  -0.97\% & +63\%  &    -4.93\% \\ \hline
        NeuralTalk   & +30\% &  -9.09\% & -  &  - \\ \hline
        Denoise      & +29\% &  -1.10\% & -  &  - \\ \hline
\textbf{geomean} & +49\% & -2.28\% & +144\% &  -9.53\% \\ \hline
\end{tabular}
\vspace{\floatMargin}
\end{table}

\begin{figure}
\centering
\includegraphics[width=0.4\textwidth]{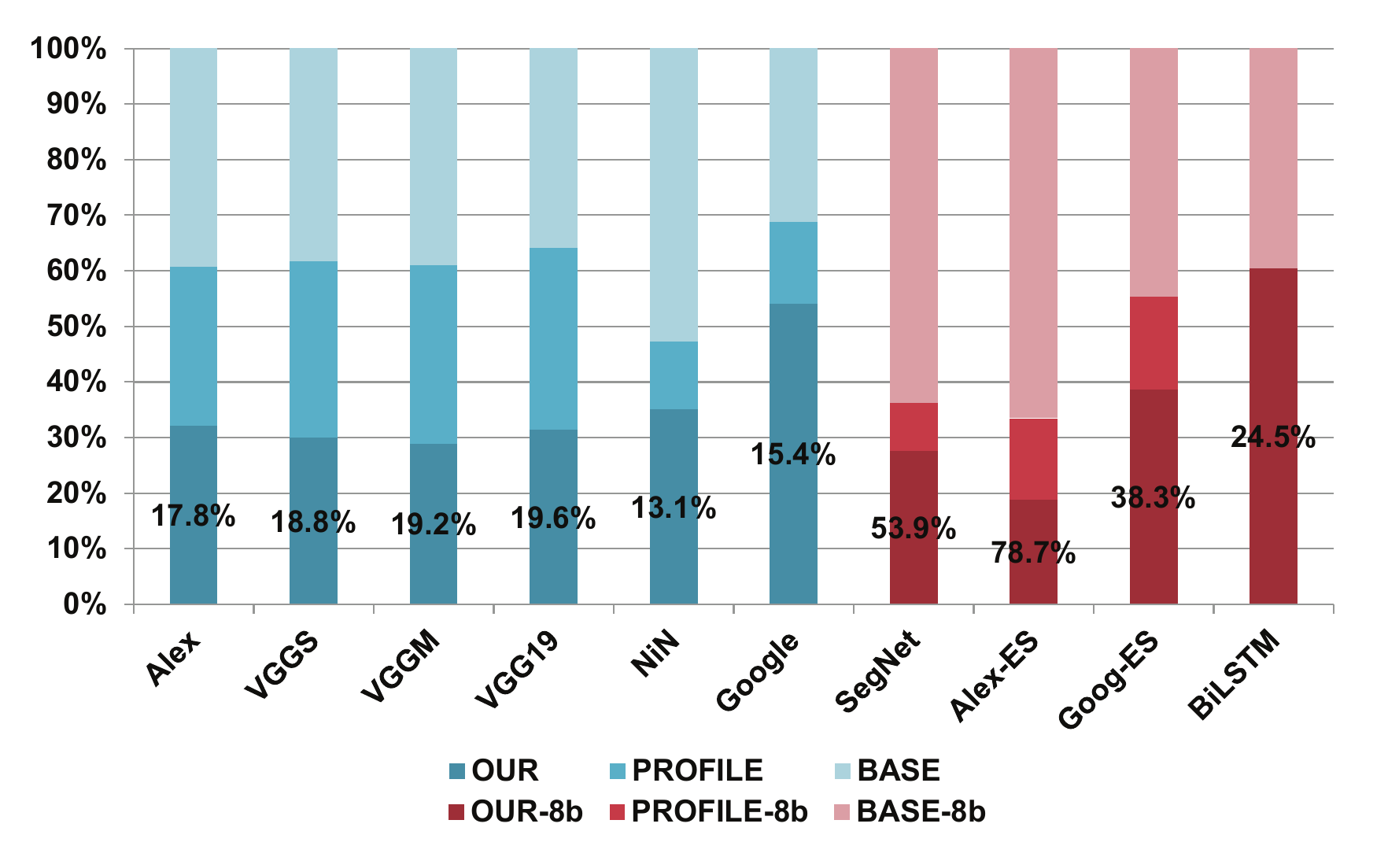}
\caption{Offchip traffic reduction with compression schemes. { Overhead for the compression metadata is reported as a percentage of the compressed traffic. \hl{The four rightmost models use per-layer precision aware 8-bit quantization.} }  }
\label{fig:compression_graph}
\vspace{\floatMargin}
\end{figure}

\begin{figure}
\centering
\includegraphics[width=0.5\textwidth]{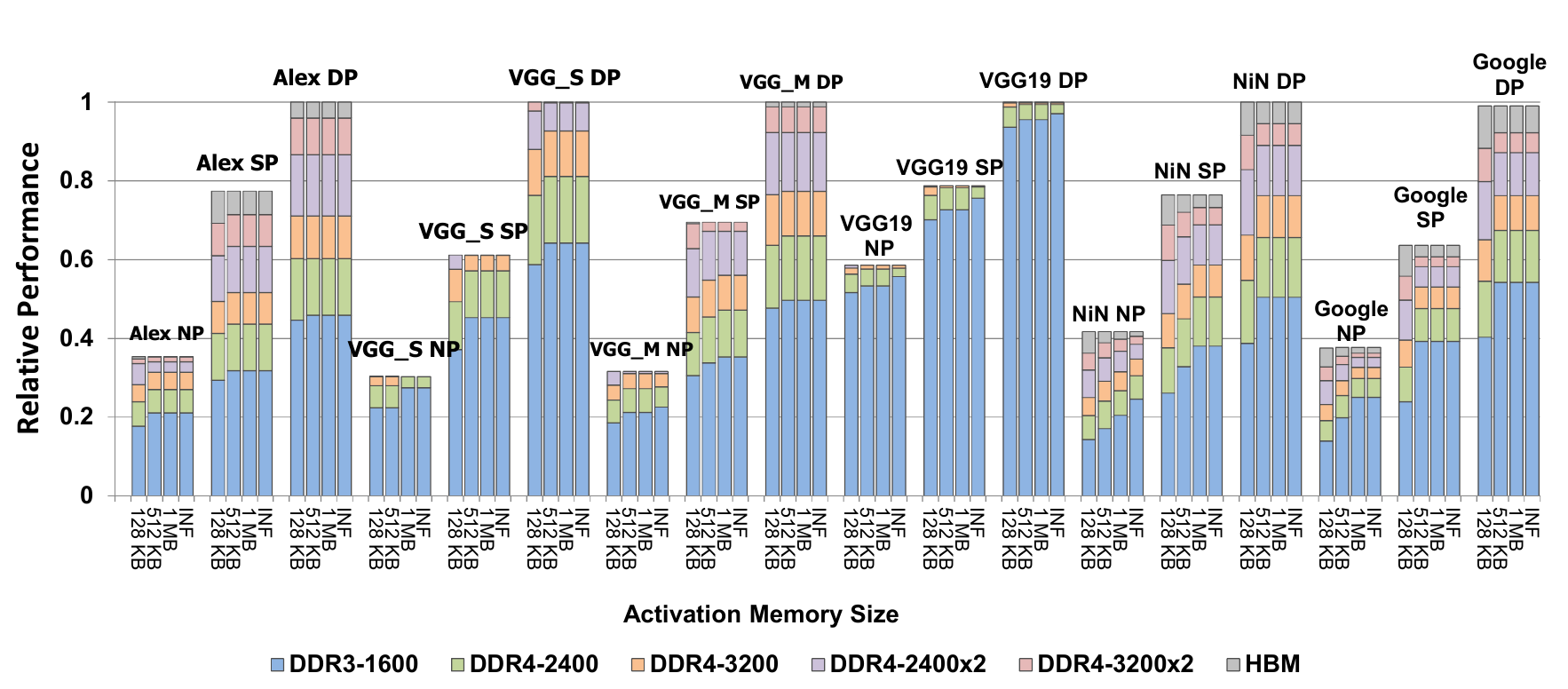}
\caption{Performance of \DSTR with different memory technologies, off-chip compression schemes, and activation memory sizes for convolutional layers.}
\label{fig:mem_graph}
\vspace{\floatMargin}
\end{figure}

\begin{table}[t]
\centering
\caption{\DSTRTITLE: \hl{Speedup over an 8-bit baseline in all layers with 8-bit quantization}}
\label{8biteval-table}
\scriptsize
\begin{tabular}{|l|l|ll}
\hline
\multicolumn{2}{|c|}{Precision Oblivious Quant.} & \multicolumn{2}{|c|}{\hl{Precision Aware Quant.}} \\\multicolumn{2}{|c|}{Iso - Peak Compute Bandwidth} & \multicolumn{2}{|c|}{\hl{Iso-Area}} \\\hline
\textbf{Name}      & \textbf{Rel. Perf.} & \multicolumn{1}{l|}{\hl{\textbf{Name}}}             & \multicolumn{1}{l|}{\hl{\textbf{Rel. Perf.}}} \\ \hline
\textbf{AlexNet}   & 1.11                & \multicolumn{1}{l|}{\hl{\textbf{AlexNet {[}26{]}}}} & \multicolumn{1}{l|}{\hl{1.62}}                \\ \hline
\textbf{NiN}       & 1.36                & \multicolumn{1}{l|}{\hl{\textbf{GoogleNet {[}26{]}}}}    & \multicolumn{1}{l|}{\hl{1.54}}                \\ \hline
\textbf{GoogleNet} & 1.28                & \multicolumn{1}{l|}{\hl{\textbf{BiLSTM}}}           & \multicolumn{1}{l|}{\hl{1.76}}                \\ \hline
\textbf{VGG\_M}    & 1.15                & \multicolumn{1}{l|}{\hl{\textbf{SegNet}}}           & \multicolumn{1}{l|}{\hl{2.50}}                \\ \hline
\textbf{VGG\_S}    & 1.22                & \multicolumn{1}{l|}{\hl{\textbf{geomean}}}          & \multicolumn{1}{l|}{\hl{\textbf{1.82}}}       \\ \hline
\textbf{VGG\_19}   & 1.44                &                                                &                                          \\ \cline{1-2}
\textbf{ResNet50}  & 1.39                &                                                &                                          \\ \cline{1-2}
\textbf{Mobilenet} & 1.27                &                                                &                                          \\ \cline{1-2}
\textbf{geomean}   & \textbf{1.27}       &                                                &                                          \\ \cline{1-2}
\end{tabular}
\end{table}

\begin{figure}
\centering
\includegraphics[width=0.45\textwidth]{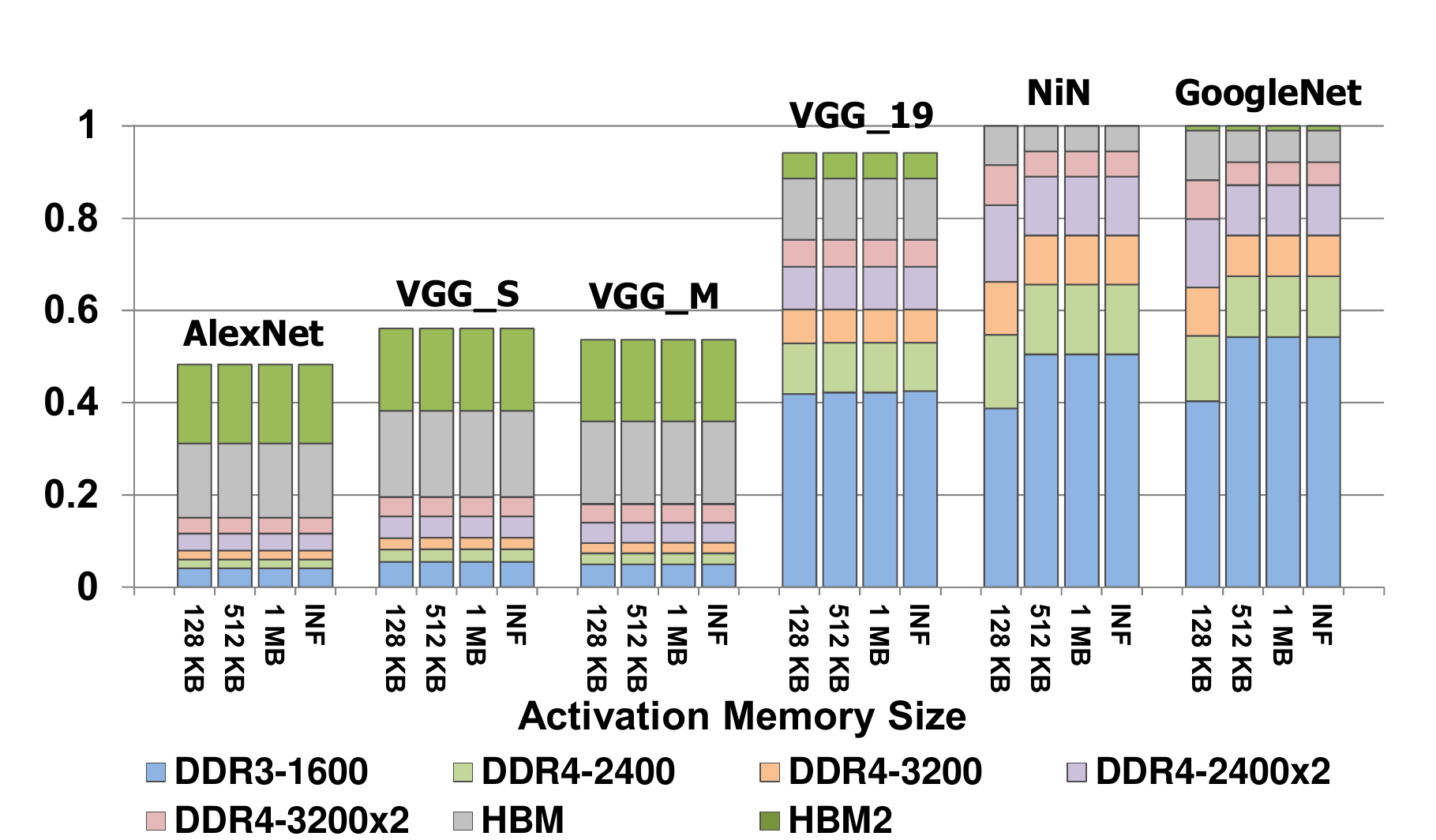}
\caption{Performance of \DSTR with different memory technologies and activation memory size for convolutional layers and fully connected layers.}
\label{fig:mem_graph_all}
\vspace{\floatMargin}
\end{figure}

\subsection{Memory Hierarchy}
We consider the effects of off-chip bandwidth and the effectiveness of \DPR for off-chip bandwidth amplification. We use 320KB of WM per tile which is sufficient so that \DSTR can read each weight only once from off-chip and per layer~\cite{kevinmemory}. Figures~\ref{fig:mem_graph} show the effect on the performance of \DSTR for different on-chip AM sizes, compression schemes ($x$-axis), and off-chip memory technologies (stacked bars). We consider three compression schemes: 1)~No Compression (NP), 2)~precision-based using profile-derived per layer precisions (SP)~\cite{ProteusICS16}, and 3)~our per group precision-based (DP). Performance is normalized to the configuration where all activations fit on chip (INF) --- the configurations used thus far. Figure~\ref{fig:mem_graph} reports relative performance for the convolutional layers only, while Figure~\ref{fig:mem_graph_all} reports relative performance for \textit{all} layers. 

Overall, the choice of external memory impacts performance significantly more than the choice of the on-chip AM. The results show that our compression scheme greatly boosts performance by reducing the impact of off-chip traffic. For the convolutional layers \DSTR when used with our compression manages to maintain most its performance benefits even with 2 channels of DDR4-3200 memory. With HBM, which would be appropriate given the peak compute capability of this configuration, the performance is within 2\% of the ideal. Off-chip memory is the bottleneck for fully-connected layers. However, using HBM or HBM2 preserves most of the overall benefits provided that our compression scheme is used. Finally, Figure~\ref{fig:compression_graph} reports the relative off-chip traffic normalized to no compression (Base). Our compression scheme reduces traffic to 55\% for GoogleNet, 60\% for BiLSTM and to less than 40\% for the remaining networks.

\subsection{\DPRTITLE Loom}
\begin{figure}
\centering
\includegraphics[width=0.4\textwidth]{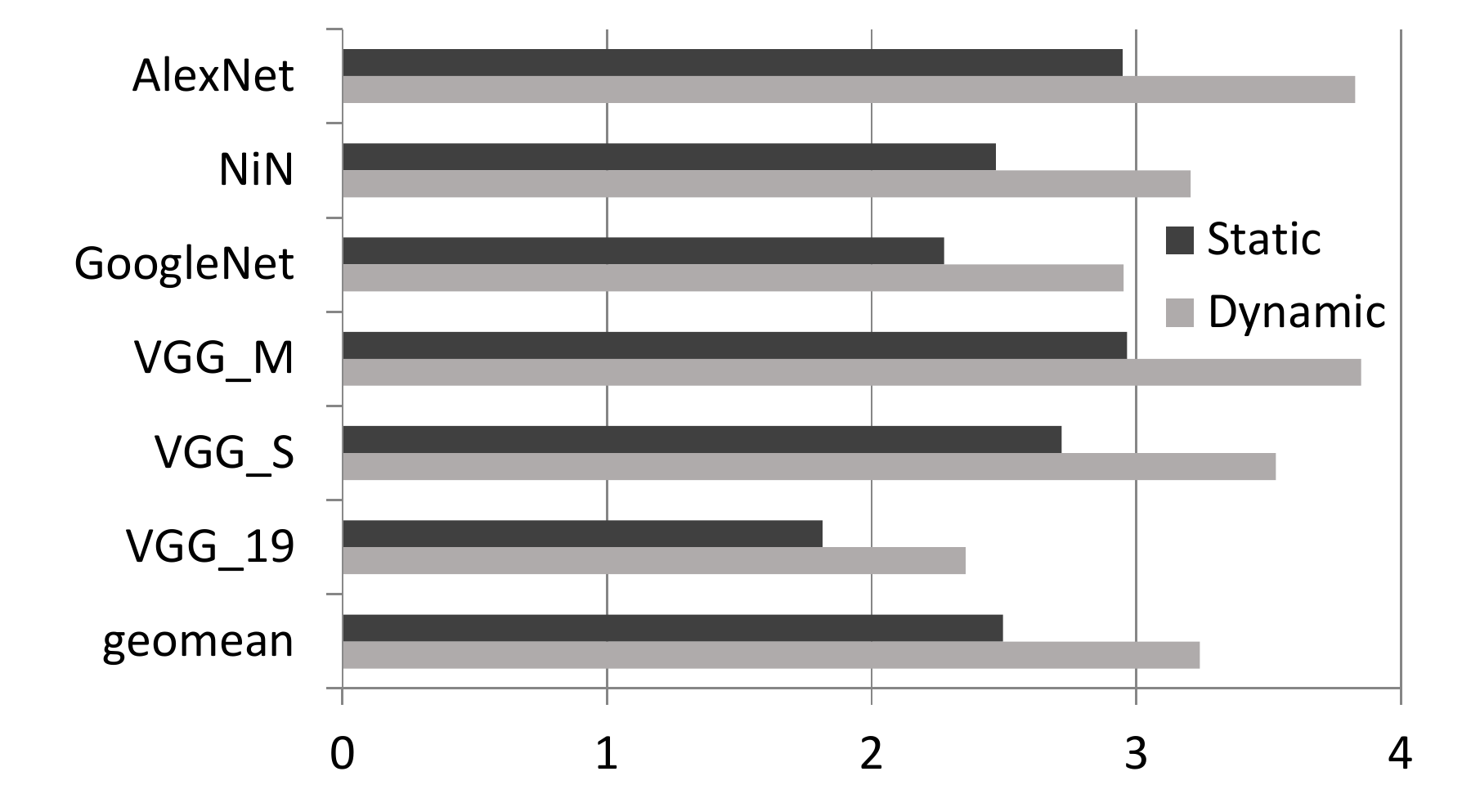}
\caption{LOOM: Comparison between static and dynamic precisions (with LPDDR4-4267)}
\label{loomeval}
\vspace{\floatMargin}
\end{figure}
Figure~\ref{loomeval} compares the performance of Loom when using dynamic vs. static per-layer precisions with a LPDDR4-4267 off-chip memory. We use the AM and WM memories suggested by Sharify \textit{et al.}~\cite{Sharify:2018:LEW:3195970.3196072}, processing 2048 weights and 256 activations concurrently bit-serially.

\subsection{8-bit Quantization}

\begin{table}[t]
\centering
\scriptsize
\caption{\hl{Area and power consumption of} \TRTEVAL \hl{normalized to} \BASE \hl{for 8-bit configuration}}
\label{table:8bit_area_power}
\begin{tabular}{|c|c|c|}
\hline
 & \textbf{\TRTEVAL (8-bit)} & \textbf{\BASE (8-bit)}\\
\hline
\hline 
\textbf{Normalized Total Area}     & $1.05\times$            & $1.00\times$     \\ \hline
\textbf{Normalized Total Compute Area}     & $1.87\times$            & $1.00\times$     \\ \hline
\textbf{Normalized Power}     & $1.64\times$            & $1.00\times$     \\ 
\hline
\end{tabular}
\vspace{\floatMargin}
\end{table}

\label{sec:quantized}
Table~\ref{8biteval-table}
\hl{reports performance for 8b quantized models. We report results for two quantization approaches. The first two columns report results with the 8-bit quantized representation used in Tensorflow}~\cite{quantizedBlog, gemmlowp}. This quantization uses 8 bits to specify arbitrary minimum and maximum limits per layer for the activations and the weights separately, and maps the 256 available 8-bit values linearly into the resulting interval. 
\hl{The second set of columns use} \textit{precision aware quantization} \hl{where quantization does not unnecessarily expand a range that can fit in less than 8b to the full 8b range. We perform an} \textbf{ iso-area comparison} \hl{ only for the last set of models. That is, we scale the baseline's compute resources to use at least as much area as that of} \DSTR; \hl{instead of 1.87x as per Table}~\ref{table:8bit_area_power} \hl{we scale the area of the baseline compute units to 2x. We do not scale the memories since this does not provide a benefit in this evaluation, as all data fits on chip. In the time alloted we were able to quantize the four models shown which include a captioning LSTM and a segmentation CNN.}

Table~\ref{table:8bit_area_power}\hl{ reports the relative area and instantaneous power of the baseline and }\DSTR \hl{when scaled to 8b. Scaling to 8b reduces area and power overheads for } \DSTR \hl{considerably since less pipelining is needed to achieve the frequency target, and quadratic scaling characteristics are diminished.}

\begin{table}[]
\scriptsize
\centering
\caption{\hl{LOOM: Speedup over an 8-bit baseline in all layers for 8-bit networks}}
\label{8biteval-table-loom}
\begin{tabular}{|l|l|}
\hline
\textbf{Name}             & \textbf{Rel. Perf.} \\ \hline
\textbf{AlexNet {[}26{]}} & 2.70                \\ \hline
\textbf{Goog {[}26{]}}    & 2.13                \\ \hline
\textbf{BiLSTM}           & 2.67                \\ \hline
\textbf{SegNet}           & 4.07                \\ \hline
\textbf{GeoMean}          & \textbf{2.81}       \\ \hline
\end{tabular}
\vspace{\floatMargin}
\end{table}

\hl{Finally,} Table~\ref{8biteval-table-loom} \hl{compares the performance of }  \DPR \hl{LOOM}~\cite{Sharify:2018:LEW:3195970.3196072} \hl{ vs. an throughput-equivalent bit-parallel baseline for the 8b precision-aware quantized models. }

\section{Related Work}
\label{sec:related}

Distributed Arithmetic (DA), which computes inner products bit-serially, has been used in the design of Digital Signal Processors~\cite{white_applications_1989}, including designs with application-specific dynamic precision detection~\cite{xan-low_power_dct-jssc2000}. DA precomputes and stores in a table all combinations of coefficients. This  is intractable for the weights of DNN as they are too many.

\textit{Pragmatic} uses a similar organization to \STR and thus \DSTR but its performance on convolutional layers depends only on the non-zero activation bits~\cite{Pragmatic}. \DPR can improve energy efficiency for Pragmatic by reducing the number of bits sent from off-chip memory and from AM to the tile. 
\DSTR represents a different area vs. efficiency vs. performance trade-off than Pragmatic. The \textit{Efficient Inference Engine} (EIE) uses weight pruning and sharing, zero activation elimination, and network retraining to drastically reduce the weight storage and communication when processing fully-connected layers~\cite{EIEISCA16}. This is an aggressive form of weight compression requiring a secondary table lookup to decode each weight. \DPR attacks activations also and thus a phenomenon that is mostly orthogonal to those EIE exploits. It may be possible to incorporate \DSTR's approach within the tile of EIE. For fully-connected layers, \TRT offers lightweight memory compression. An appropriately configured EIE will may outperform \TRT for fully-connected layers, provided that the network is pruned and retrained. { However, while pruning in earlier CNNs resulted in sparsity in the 80\% to 90\% range, for more recent networks sparsity is lower, in the 50\% to 60\% range}~\cite{cvpr_2017_yang_energy,SkimCaffePaper}.

Bit-Fusion takes advantage of per layer precisions using a spatial design instead of the bit-serial design of Stripes~\cite{bitfusion}. \DSTR adapts precisions at a finer granularity and dynamically. Bit-Fusion can benefit from the off-chip compression offered by \DPR. 

Advances at the algorithmic level would impact \DSTR and \TRT as well or may even render them obsolete. For example, work on using binary weights~\cite{courbariaux2015binaryconnect} would obviate the need for an accelerator whose performance scales with weight precision.

\hl{New datatypes such as Intel's Flexpoint}~\cite{flexpoint} \hl{or Tensorflow's \textit{bfloat16}}~\cite{tensorflow2015-whitepaper} \hl{have been introduced as lower precision alternatives to 32-bit floating point for use in deep neural networks. They provide a wider range than fixed-point representations and target primarily training.  They are }\textit{statically} \hl{sized data representations and require 7, 8 or 16 bits of mantissa, and 8 bits of exponent regardless of the magnitude of the value stored. The models will still exhibit a lopsided value distribution and will natural exhibit variable precision requirements per layer. Accordingly, } \DPR \hl{ can be applied to both the mantissa and exponent  to adjust the least significant bits used after profiling has been used to trim the precision needed per layer.}

\hl{Quantization to lower precisions has been used to reduce the memory and computation cost in neural networks. For example, Google's Tensor Processing Unit (TPU) uses quantization to represent values using 8 bits}~\cite{TPUISCA17} \hl{ to support TensorFlow}~\cite{tensorflow2015-whitepaper}. \hl{Nvidia TensorRT also supports quantization to 8-bits}~\cite{nvidia_8bit}. \hl{However, quantization is not without its drawbacks. Although a fixed precision may be suitable for certain networks, using a fixed quantization scheme may not be optimal/sufficient for all problems, e.g., those using RNNs}~\cite{nvidia_8bit}\hl{or computational imaging tasks performing per-pixel prediction, e.g.,}~\cite{VDSR,IRCNN}\hl{ which process raw sensor data of 12b or more, or even some image classification tasks and speech processing}~\cite{DBLP:journals/corr/abs-1712-05877}\hl{. Therefore, flexible precision support is important to generalize across different applications.}\hl{While quantization attenuates some of the expected benefits with }\DPR \hl{the underlying phenomenon that }\DPR \hl{ exploits persists: there will be few high values. This is the reason why it benefits even 8b quantized image networks. Further, }\DSTR \hl{ and LOOM offer benefits even for those networks that can be quantized further even for non powers of two precisions, e.g.,}~\cite{DBLP:conf/eccv/ParkYV18} \hl{ without requiring all networks to be quantized to obtain benefits}. 
Park et al., present an outlier aware accelerator design where a set of limited high-precision execution units are used to process those neurons that require them whereas the bulk of computations happen in lower precision units~\cite{outlier}. For this purpose, the weights are quantized and partitioned statically. \DPR works with any network whether quantized or not and adapts the precisions of both weights and activations. For this reason it could be beneficial even for the above reference accelerator. 

\section{Conclusion}
\label{sec:theend}

We believe that dynamic precision reduction opens up several directions for future work including how to combine with other accelerator engines, how it can boost the effectiveness of algorithms for pruning or for precision reduction or quantization of weights and of activations, and whether it can be exploited to accelerate training as well. The accelerators we proposed \textit{do not require} any changes to the input network to deliver benefits. However, they \textit{do reward} any advances in precision reduction   including linear quantization and thus if deployed will provide an incentive for further innovation towards networks of extremely low precision.

\bibliographystyle{ieeetr}
\bibliography{ref}

\end{document}